%% file: iclr2027_conference.tex
\documentclass{article} 
\usepackage{iclr2027_conference,times}

\input{math_commands.tex}

\usepackage{hyperref}
\usepackage{url}
\usepackage{algorithm}
\usepackage{algpseudocode}
\usepackage{booktabs}
\usepackage{graphicx}
\title{DT-JEPA: Learning to Predict Future Representations for Decision-Making}

\author{
Changxu Ni$^{1,*}$ \quad
Yiyang Ni$^{2,*}$ \quad
Jun Li$^{1,\dagger}$ \quad
Feiran You$^{3,*,\dagger}$ \quad
Hongyang Du$^{3,\dagger}$ \quad
Haitao Zhao$^{4}$ \quad
Shi Jin$^{1}$
\\[0.7em]
$^{1}$Southeast University
\\
$^{2}$Jiangsu Second Normal University
\\
$^{3}$The University of Hong Kong
\\
$^{4}$National University of Defense Technology
}

\iclrfinalcopy 
\begin{document}

\maketitle
\begingroup
\renewcommand{\thefootnote}{\fnsymbol{footnote}}
\footnotetext[1]{Equal contribution.}
\footnotetext[2]{Corresponding authors.}
\endgroup
\begin{abstract}
Decision Transformer (DT) has demonstrated strong performance in various offline reinforcement learning (RL) tasks by conditioning action generation on expected future returns and historical trajectories. However, treating expected future returns as the condition provides an incomplete description of the future, as trajectories with different state transitions and behaviors may share similar returns. Moreover, expected future returns inherently depend on task-specific reward functions, which makes the policy learning reliant on reward-labeled datasets and restricts its scalability to tasks with similar but unseen rewards. Recently, Joint Embedding Predictive Architectures (JEPA) has emerged as a promising paradigm of self-supervised learning-based world models, which learns predictive representations of future dynamics in latent space rather than reconstructing future states in the original space. Inspired by JEPA, we propose DT-JEPA, an offline pretraining and online finetuning framework that learns general knowledge from reward-free offline trajectories and achieve excellent adaptation to the unseen tasks. Instead of predicting a single deterministic future embedding, we design a novel probabilistic predictor to model the
conditional distribution of future representations in the latent space, thereby
capturing both predictive information and uncertainty about future
dynamics. The predicted future representation is then utilized to guide DT's action generation. To generalize to tasks with unseen reward objectives, we further develop a critic-guided fine-tuning mechanism to efficiently adapt the pretrained policy to unseen reward functions with limited online interactions. Theoretically, we show that 1) the JEPA objective seeks to maximize a variational lower bound on the mutual information between historical representations and future trajectories; and 2) predictive future-conditioned policy learning admits a finite-sample performance bound with an $\mathcal{O}(N^{-1/4})$ estimation term. Experiments on D4RL benchmarks demonstrate that DT-JEPA achieves efficient adaptation across varying offline RL tasks, outperforming traditional DT-based and other offline RL methods.
\end{abstract}

\section{Introduction}

Complex decision-making domains such as robotic control, autonomous driving, and embodied intelligence often involve large-scale offline datasets collected across diverse tasks and policies. Learning general decision knowledge from such heterogeneous data and efficiently adapting it to new downstream tasks is becoming increasingly important~\citep{levine2020offline}. Motivated by the success of pretraining--finetuning paradigms in computer vision~\citep{chen2020simclr,he2022mae} and natural language processing~\citep{devlin2019bert,radford2018improving}, reinforcement learning (RL) research has recently focused on leveraging offline trajectories to learn reusable policies and representations that can generalize across tasks. Offline RL provides a natural foundation for this direction, as it enables policy learning entirely from previously collected experience without requiring additional interaction with the environment. More recently, inspired by the strong sequence modeling capability of Transformers~\citep{vaswani2017attention}, subsequent studies have reformulated offline policy learning as a conditional sequence modeling problem~\citep{chen2021decision}, offering a new perspective for offline RL.

Within the CSM formulation, policy inference is conditioned on a specific target together with the historical context. Following this paradigm, the representative Decision Transformer (DT) conditions action generation on desired future returns and has demonstrated strong performance across a wide range of offline RL tasks. Despite its effectiveness, return-oriented CSM methods still suffer from several limitations. On the one hand, expected future returns provide an ambiguous description of future behaviors, since distinct future trajectories with different state transitions and action patterns may share similar cumulative returns. On the other hand, return conditioning inherently relies on task-specific reward functions, restricting policy pretraining to reward-annotated trajectories and thereby limiting scalability and transferability. To alleviate these limitations, Generalized DT (GDT) extends return conditioning to generalized hindsight statistics extracted from future trajectories~\citep{furuta2022generalized}, while Pretrained DT (PDT) encodes future trajectory segments into embeddings for future-conditioned action generation~\citep{xie2023future}. However, these approaches still rely on realized future trajectory information during pretraining and require appropriate future conditions to be specified or selected at deployment. This raises an important question: \textit{How can we explicitly learn the distribution of future dynamics  in the latent space solely from observable history, while ensuring
the predicted representations preserve sufficient information about the
future to guide DT's action generation?}

Recent advances in offline RL have begun to explore policy learning under reward-free and unsupervised pretraining paradigms~\citep{touati2021learning,park2024hilp,bagatella2026tdjepa}. Since the transition dynamics $p(s' \mid s,a)$ characterize how the environment evolves under different actions, agents can still extract substantial decision-relevant knowledge from offline trajectories without explicit reward supervision. World models provide a natural framework for capturing such dynamics by learning predictive models of environment transitions~\citep{ha2018world,hafner2020dream}. However, conventional generative world-modeling approaches often rely on observation reconstruction or generation objectives, which may require modeling fine-grained details that are not directly relevant to downstream decision making. Recently, Joint-Embedding Predictive Architectures (JEPA) have emerged as a promising alternative for predictive representation learning~\citep{lecun2022path,assran2023ijepa,bardes2024vjepa}. Rather than reconstructing future observations in the original input space, JEPA predicts their latent representations from observed context, encouraging the model to capture abstract information that is predictive of future dynamics.

Inspired by this paradigm, we propose DT-JEPA, a generalizable offline pretraining and online fine-tuning framework that learns transferable decision knowledge from reward-free offline trajectories and efficiently adapts to unseen downstream tasks. First, to capture the predictive information and uncertainty
about future dynamics, we design a novel probabilistic predictor to model the conditional distribution of future representations in the latent space. Second, we incorporate the predicted future representation into DT through adaptive layer normalization (AdaLN)~\citep{peebles2023dit}, dynamically modulating historical representations based on the future-predictive information to guide DT's action generation. Finally, to adapt the pretrained policy to a new task with an unseen reward function, we develop a critic-guided online fine-tuning mechanism. The DT-JEPA policy and critic are jointly optimized on newly collected trajectories, where the critic estimates action values under the target reward and provides Q-value guidance for adapting the policy toward the new task objective.

Theoretically, we show that 1) JEPA objective function maximize a variational lower bound on mutual information between historical representations and future trajectories; and 2) predictive future-conditioned policy learning admits a finite-sample performance bound with an $\mathcal{O}(N^{-1/4})$ estimation term. Empirically, we evaluate DT-JEPA on the standard D4RL benchmark (Fu et al., 2020). Experimental results demonstrate that DT-JEPA consistently achieves superior performance over existing DT variants and competitive offline RL methods. In particular, DT-JEPA shows strong generalization to the unseen tasks with unseen reward functions, achieving efficient online adaptation with limited interactions. 
\section{Preliminaries}
\label{gen_inst}

\subsection{Offline Reinforcement Learning and Decision Transformer}

We consider a Markov decision process (MDP)
$\mathcal{M}=(\mathcal{S},\mathcal{A},P,r,\gamma,\rho_0)$,
where $\mathcal{S}$ and $\mathcal{A}$ denote the state and action spaces,
$P(s'|s,a)$ is the transition dynamics,
$r:\mathcal{S}\times\mathcal{A}\rightarrow\mathbb{R}$ is the reward function,
$\gamma\in[0,1)$ is the discount factor, and $\rho_0$ is the initial-state
distribution~\citep{sutton2018reinforcement}. A policy $\pi(a|s)$ aims to maximize the expected discounted
return $J(\pi)
    =
    \mathbb{E}_{\pi}
    \left[
        \sum_{t}
        \gamma^t r(s_t,a_t)
    \right].$
The corresponding action-value function is defined as
$Q^{\pi}(s_t,a_t)
    =
    \mathbb{E}_{\pi}
    \left[
        \sum_{k}
        \gamma^k
        r(s_{t+k},a_{t+k})
        \,\middle|\,
        s_t,a_t
    \right].$
In offline reinforcement learning,
the agent cannot interact with the environment during training.
Instead, learning is performed exclusively from a fixed dataset
$\mathcal{D}$ collected beforehand by one or multiple behavior policies~\citep{levine2020offline}.
DT reformulates offline RL
as a conditional sequence modeling problem. For each trajectory, DT augments
the state--action sequence with the return-to-go (RTG)
$R_t
    =
    \sum_{t'=t}^{T} r_{t'},$
which summarizes the cumulative future reward from the current timestep.
The resulting sequence is written as
    $\hat{\tau}
    =
    (R_1,s_1,a_1,
     R_2,s_2,a_2,
     \ldots,
     R_T,s_T,a_T).$
After tokenization and positional encoding, the sequence is processed by a
causal Transformer that predicts the current action from previously observed
trajectory tokens and the desired RTG. In probabilistic form, the policy can
be expressed as
$\pi_{\theta}
    \left(
        a_t
        \mid
        \hat{\tau}_{1:t-1},s_t,R_t
    \right).$
The standard DT objective is therefore an action prediction loss over the
offline trajectories,
\begin{equation}
    \mathcal{L}_{\mathrm{DT}}
    =
    \mathbb{E}_{\hat{\tau}\sim\hat{\mathcal{D}}}
    \left[
        \sum_{t=1}^{T}
        -\log
        \pi_{\theta}
        \left(
            a_t
            \mid
            \hat{\tau}_{1:t-1},s_t,R_t
        \right)
    \right],
\end{equation}
where $\hat{\mathcal{D}}$ denotes the reward-augmented trajectory dataset.
Thus, unlike conventional value-based offline RL, DT learns a policy primarily
through supervised trajectory modeling, with RTG serving as an explicit
conditioning signal describing the desired future outcome.

\subsection{Joint-Embedding Predictive Architecture}

The Joint Embedding Prediction Architecture (JEPA) is a self-supervised representation learning framework based on latent-space prediction~\citep{lecun2022path,assran2023ijepa}. Its core idea is not to directly reconstruct the original observed data, but rather to predict the latent-space embeddings of missing target data. Given an input sample, JEPA constructs a context view $x_C$ and a target
view $x_T$, which are processed by a context encoder $f_{\theta}$ and a
target encoder $f_{\theta'}$, respectively. This produces the latent
representations $\mathbf{z}_C=f_{\theta}(x_C)$ and
$\mathbf{z}_T=f_{\theta'}(x_T)$.
The parameters of the target encoder $f_{\theta'}$ are updated using an Exponential Moving Average (EMA) of the context encoder’s parameters, i.e.,
$\theta' \leftarrow \alpha\theta' +(1-\alpha)\theta$, where $\alpha\in[0,1)$ is the momentum coefficient~\citep{grill2020byol}.
Building on this, JEPA uses a predictor $g_{\phi}$ to predict the target latent representation based on the context representation $\mathbf{z}_C$ and target-related auxiliary information $\xi_T$, i.e.,
$\hat{\mathbf{z}}_T=g_{\phi}(\mathbf{z}_C,\xi_T)$.
The model is trained by minimizing the distance between the predicted representation and the target representation:
\begin{equation}
\mathcal{L}_{\mathrm{JEPA}}
=
\left\|
g_{\phi}
\left(
f_{\theta}(x_C),\xi_T
\right)
-
f_{\theta'}(x_T)
\right\|_2^2.
\end{equation}
Thus, through the above method, JEPA ensures that the latent representation $\mathbf{z}_C$ retains information that is relevant to and predictive of the target $x_T$.

\subsection{Adaptive Layer Normalization}

AdaLN is an adaptive normalization mechanism that injects conditional information into the feature representations of neural networks~\citep{peebles2023dit}. AdaLN dynamically generates corresponding affine transformation parameters based on an external conditional variable $z$. Specifically, it first uses a lightweight mapping network to convert the conditional representation into scaling coefficients and a bias term, i.e.,
$\gamma(z)=\operatorname{MLP}_{\gamma}(z), \beta(z)=\operatorname{MLP}_{\beta}(z)$.
Subsequently, given an input feature $x$, AdaLN conditionally modulates its normalized representation, i.e.,
\begin{equation}
\operatorname{AdaLN}(x,z)
=
\gamma(z)\odot
\frac{x-\mu(x)}
{\sqrt{\sigma^2(x)+\epsilon}}
+
\beta(z),
\end{equation}
where $\mu(x)$ and $\sigma^2(x)$ represent the mean and variance along the feature dimension, respectively, $\epsilon$ is a constant ensuring numerical stability, and $\odot$ denotes element-wise multiplication. By allowing the scaling and bias parameters to adaptively vary with the conditional variable $z$, AdaLN is able to incorporate conditional information into the feature representation without explicitly adding the conditional variable to the input sequence.
\section{Method}
In this section, we present DT-JEPA, an offline pretraining and online finetuning-based predictive sequence-modeling
framework for generalizable decision making. During the pretraining phase, a variational JEPA model is designed to infer future representations from multi-task offline
data without reward supervision. Subsequently, the predicted future latent guides DT's action generation by effectively modulating transformer's hidden representations. During the finetuning phase, task-specific critics is designed to efficiently adapt the pretrained policy toward unseen reward objectives with
limited environment interaction. Theoretically,
we analyze the effectiveness of the  predictive representation learning objective and derive a finite-sample performance
bound for the future-conditioned policy 
learning.

\subsection{Motivation}
\begin{figure*}[t]
\centering
\includegraphics[width=1\textwidth]{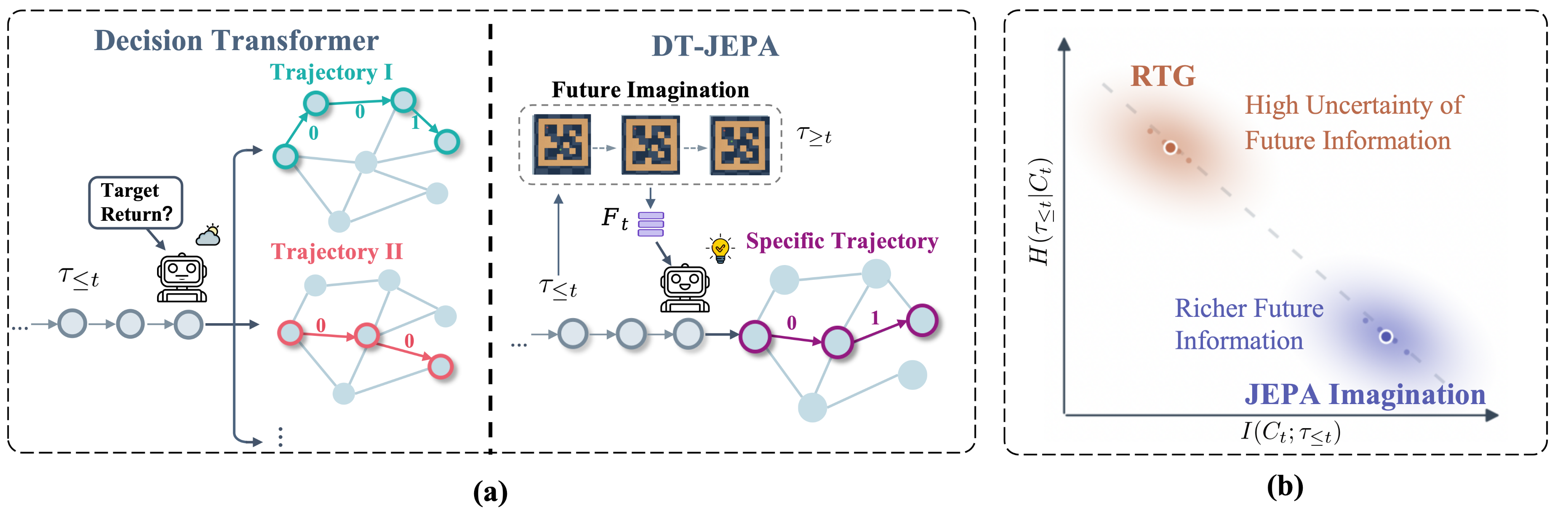}
\caption{(a) RTG summarizes future with a ambiguous target return, while DT-JEPA instead predicts latent future dynamics from historical context and effectively guide current decision making.
(b) From an information-theoretic perspective, a more informative future condition reduces uncertainty about future trajectories and preserves richer future information.}
\label{fig:training_framework}
\end{figure*}
The DT uses RTG, i.e. $R_t=\sum_{i=t}^T r_t$ as the condition to guide the generation of the current action. Let $\tau_{\rm \ge t}$ denote the future trajectory that may arise after the current time step. It is obvious that significant 
different $\tau^A_{\rm \ge t}$ and $\tau^B_{\rm \ge t}$ which exhibit different state transitions and actions may correspond to very similar returns $R^A_t$ and $R^B_t$. 
From an information-theoretic perspective, we can explain this through conditional entropy and mutual information. For any conditional variable $C_t$ used in the current decision, its mutual information with the future trajectory $I(C_t;\tau_{\ge t})$ can be expressed as $I(C_t;\tau_{\ge t}) = H(\tau_{\ge t}) - H(\tau_{\tau_{\ge t}}\mid C_t)$. Here, $H(\tau_{\ge t})$ represents the entropy of the future trajectory $\tau_{\tau_{\ge t}}$ and is used to characterize the uncertainty of its overall distribution; $H(\tau_{\ge t}\mid C_t)$ represents the conditional entropy of the future trajectory given the conditional variable $C_t$, which measures the uncertainty that remains in the distribution of the future trajectory once this condition is known. As shown in the Figure 1, the smaller $H(\tau_{\ge t}\mid C_t)$ is, the more effectively the conditional variable $C_t$ reduces uncertainty about the future, thereby increasing the mutual information $I(C_t;\tau_{\ge t})$. In traditional DT, RTG is used as the conditional variable. Since different future trajectories may correspond to similar RTG values, $H(\tau_{\ge t}\mid R_t)$ is relatively large, thereby reducing the mutual information $I(R_t;\tau_{\ge t})$ between the conditional variable RTG and the future trajectory.
To address this issue, drawing on the concept of the JEPA world model, we seek to learn $\hat{F}_t$ as the future latent representation predicted by JEPA based on historical trajectories, which is expected to preserve richer predictive information about future trajectories beyond what can be represented by a scalar RTG. 

\subsection{Variational Joint-Embedding Future Conditioning}
To predict a more effective future representation as the conditional variable for the DT's action generation, we constructing a JEPA-based future representation learning mechanism. Specifically, the observable history of DT at each time step is defined as
$\tau_{t-K+1: t}
=
\left(
s_{t-K+1:t},
a_{t-K+1:t-1}
\right),$
which is padded to the fixed length when the available history length is less than $K$.
Based on the $\tau_{t-K+1:t}$, we  construct a historical trajectory encoder
$f_{\theta}$, which maps $\tau_{t-K+1:t}$ to the historical representation $\mathbf{H}_{t}$, i.e.,
\begin{equation}
\mathbf{H}_{t}=f_{\theta}
\left(
\tau_{t-K+1:t}
\right).
\end{equation}

Moreover, based on the future trajectory  $\tau_{t:t+K-1}$ from the offline dataset $\mathcal{D}$, we construct a target future encoder
$f_{\theta'}$ to model the target future representation $\mathbf{F}_{t}$, i.e.,
\begin{equation}
\mathbf{F}_{t}=f_{\theta'}
\left(
\tau_{t:t+K-1}
\right).
\end{equation}
where $\theta'$ is updated from $\theta$ via EMA.
Based on $\mathbf{H}_{t}$ obtained from the historical trajectory encoder, we further construct a probabilistic predictor
$q_{\eta}
\left(
\cdot
\mid
\mathbf{H}_{t}
\right)
=
\mathcal{N}
\left(
\hat{\boldsymbol{\mu}}_{t}^{F},
\operatorname{diag}
\left(
(\hat{\boldsymbol{\sigma}}_{t}^{P})^{2}
\right)
\right),$
where
$\hat{\boldsymbol{\mu}}_{t}^{P}$ and $\hat{\boldsymbol{\sigma}}_{t}^{P}$
denote the mean and variance of the predicted future representation. Based on $\mathbf{H}_t$, we then sample predicted future representation latent as
$\hat{\mathbf{F}}_{t}\sim q_{\eta}
\left(
\cdot
\mid
\mathbf{H}_{t}
\right). $

To encourage the predicted future representation $\hat{\mathbf F}_t$ to capture informative patterns of future trajectory evolution, we seek to encourage the historical representation $\mathbf H_t$ to
retain information that is predictive of the future trajectory $\tau_{\ge t}$. Accordingly, we derive a variational lower bound on the conditional mutual information $I(\tau_{\ge t};\mathbf H_t)$, which further motivates the learning objective of the JEPA module.

\textbf{Theorem 1.}
\textit{The vaiational lower bound of mutual information $I(\tau_{\ge t};\mathbf H_t)$ satisfies}:
\begin{equation}
I(\tau_{\ge t};\mathbf H_t)
\geq
\mathbb E_{p(\mathbf  F_t,\mathbf H_t)}
\left[
\log q_{\eta}(\mathbf F_t\mid\mathbf H_t)
\right]
+
H(\mathbf F_t),
\label{eq:mi_lower_bound}
\end{equation}
\textit{where $p(\mathbf F_t,\mathbf H_t)$ denotes the joint
distribution of the historical and future representations, $H(\mathbf F_t)$ denotes the  entropy of the future representation $\mathbf F_t$}.

The proof of the Theorem 1 are provided in Appendix B.1. Since $H(\mathbf F_t)$ depends only on the target encoder, and the target encoder is updated gradually via EMA, it changes relatively slowly during training. Therefore, based on the variational lower bound in Theorem 1, we design the loss function for the JEPA module as
\begin{align}
    \mathcal{L}_{\mathrm{JEPA}}
=\;
\mathbb{E}_{
(\tau_{t-K+1:t},\tau_{t:t+K-1})
\sim\mathcal{D}}
&\left[
-\log
q_{\eta}
\left(
\mathbf{F}_t
\mid
\mathbf{H}_t
\right)
\right]
\end{align}
where $\mathcal{D}$ denotes the trajectory dataset,
and $\theta'$ employs a stop-gradient mechanism.
\begin{figure*}[t]
\centering
\includegraphics[width=1\textwidth]{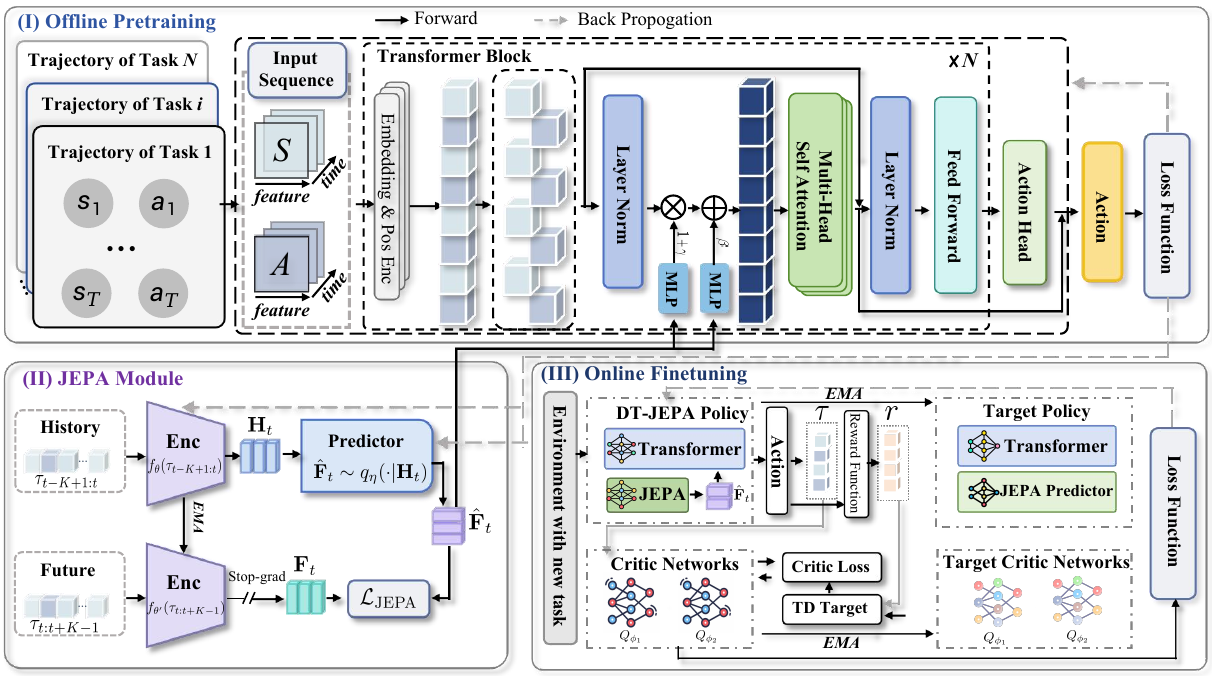}
\caption{\textbf{Overview of the proposed DT-JEPA framework}. DT-JEPA learns to make decisions using predicted future-dynamics information. During pretraining, a probabilistic JEPA predictor generates latent future representations to modulate DT through AdaLN. During online fine-tuning, specific critics provide value guidance to adapt the pretrained policy to unseen reward objectives.}
\label{fig:training_framework}
\end{figure*}
\subsection{AdaLN-based Latent Conditioned Action Generation}

To effectively utilize the predicted future representation
$\hat{\mathbf F}_t$ for guiding the DT's action generation, we introduce an AdaLN-based latent modulation mechanism. Given the historical trajectory $\tau_{t-K+1:t}$, DT first extracts its hidden representation $\mathbf h_t$. The predicted future latent $\hat{\mathbf F}_t$ is then used as a conditioning signal to adaptively modulate $\mathbf h_t$. Specifically, two lightweight MLPs map $\hat{\mathbf F}_t$ to dimension-wise scaling and shifting parameters, i.e.,
\begin{equation}
\boldsymbol{\gamma}_t
=
\mathrm{MLP}_{\gamma}
\left(
\hat{\mathbf F}_t
\right),
\quad
\boldsymbol{\beta}_t
=
\mathrm{MLP}_{\beta}
\left(
\hat{\mathbf F}_t
\right).
\end{equation}
The historical representation is normalized and subsequently modulated as
\begin{equation}
\tilde{\mathbf h}_t
=
\left(
1+\boldsymbol{\gamma}_t
\right)
\odot
\operatorname{LayerNorm}
\left(
\mathbf h_t
\right)
+
\boldsymbol{\beta}_t,
\end{equation}
where $\odot$ denotes the Hadamard product. Through the feature-wise scaling and shifting operations, the predicted
future latent variable is effectively incorporated into the historical
representation. Finally, the modulated decision representation $\tilde{\mathbf h}_t$ is used to predict the action $\hat{\mathbf a}_t$ at the current timestep.
\subsection{Reward-Free Pretraining and Reward-Oriented Finetuning}

During the pretraining stage, we first train the policy using a standard behavior cloning (BC) objective on large-scale trajectory data of multiple tasks~\citep{chen2021decision}. Specifically, the BC loss is defined as
\begin{equation}
\mathcal L_{\mathrm{BC}}
=
\mathbb E_{
\tau_{t-K+1:t}\sim\mathcal{D},\,\,
\hat{\mathbf F}_{t}\sim q_{\eta}
}
\left[
\sum_{m=t-K+1}^{t}
\left(
a_m-\hat a_m
\right)^2
\right],
\end{equation}
where $\hat a_m=\phi(\tau_{t-K+1:t},\hat{\mathbf F}_t)$ denotes the action predicted by DT, and $\phi$ denotes the policy network.
The overall objective during the pretraining stage is formulated as
\begin{equation}
\mathcal L(\phi,\theta,\eta)
=
\mathcal L_{\mathrm{BC}}
+
\lambda_J \mathcal L_{\mathrm{JEPA}},
\end{equation}
where $\lambda_J$ is a hyperparameter that balances the BC loss and the JEPA prediction loss.

To characterize whether the predictive-conditioned policy learned from
a finite offline dataset can approximate its population counterpart, we
further provide a finite-sample performance analysis. 

\textbf{Theorem 2.}
\label{thm:pfc_sample_complexity}
\textit{
Consider a finite-horizon MDP with horizon $H$ and rewards in $[0,1]$.
For a given predictor $q_\eta(F\mid x)$, let
$\bar{\pi}_\eta^{\rm P}$ and $\bar{\pi}_{\eta,\hat{\phi}}$
denote the population and learned marginal predictive-conditioned
policies. Under bounded occupancy mismatch $C_F$, finite policy class
$\Pi_F$, bounded log-likelihood variation $c_F$, and approximation
error $\epsilon_{\rm app}^{F}$, with probability at least $1-\delta$,
}
\begin{equation}
J(\bar{\pi}_\eta^{\rm P})
-
J(\bar{\pi}_{\eta,\hat{\phi}})
\leq
\mathcal O\!\left(
C_F H^2
\left[
\sqrt{c_F}
\left(
\frac{\log(|\Pi_F|/\delta)}{N}
\right)^{1/4}
+
\sqrt{\epsilon_{\rm app}^{F}}
\right]
\right).
\end{equation}

\textit{Proof:} See the Appendix B.2. Theorem~2 shows that, under a given future
predictor, the learned predictive-conditioned policy approaches its
population counterpart as the amount of offline data increases, up to
the approximation error. This result supports learning a transferable predictive-conditioned
policy from large-scale reward-free trajectories.
Building on the pretrained policy, we further adapt the policy to a new task $j$ by introducing task-specific critics that guide policy optimization toward higher expected long-term return under the new task objective~\citep{zheng2022online}. Specifically, the pretrained policy first interacts with the environment of task $j$ to collect a set of trajectories
\(
\tau^j=(s_1^j,a_1^j,\ldots,s_T^j,a_T^j)
\), which are used to construct the specific dataset $\mathcal D^j$. The trajectories with relatively high cumulative returns in $\mathcal D^j$ is further used to form a high-return subset, denoted by $\hat{\mathcal D}^j$.
We then construct two task-specific critic networks
$Q_{\omega_1^j}^j(s_t^j,a_t^j)$ and
$Q_{\omega_2^j}^j(s_t^j,a_t^j)$
to estimate the expected discounted cumulative return obtained by executing action $a_t^j$ at state $s_t^j$ under task $j$~\citep{fujimoto2018td3}.
For the two critic networks and the DT policy network, we respectively maintain EMA target networks parameterized by
$\bar{\omega}_1^j$, $\bar{\omega}_2^j$, and $\bar{\phi}$.
The temporal-difference target is defined as
\begin{equation}
\hat{Q}_m^j
=
\sum_{k=m}^{t-1}
\gamma^{k-m} r_k^j
+
\gamma^{t-m}
\min_{d\in\{1,2\}}
Q_{\bar{\omega}_d^j}^j
\left(
s_t^j,
\hat a_t^j
\right),
\end{equation}
where $\gamma$ denotes the discount factor and $r_k^j$ denotes the immediate reward at timestep $k$ under task $j$.
Thus, $\omega^j_d$ is optimized by the following loss function, i.e.,
\begin{equation}
\mathcal L(\omega^j_d)
=
\mathbb E_{\tau^j_{t-K+1:t}\sim\hat{\mathcal D^j},\mathbf{F}_{t}^{j}\sim f_{\theta},\hat{a}^j_t\sim\bar\phi}
\left[\sum_{m=t-K+1}^{t-1}
\left(
Q^j_{\omega_d}
\left(
s_m^j,a_m^j
\right)
-
\hat{Q}_m^j
\right)^2
\right].
\end{equation}

Consequently, based on the actions $\hat a_t^j$ generated by DT under the task $j$, we further utilize the critic network to evaluate  long-term values. Specifically, we define the value-guided policy loss as
\begin{equation}
\mathcal L_Q^j
=
-
\mathbb E_{
\tau_{t-K+1:t}^j\sim\hat{\mathcal D}^j,\,
\mathbf F_t^j\sim f_\theta^j
}
\left[
\sum_{m=t-K+1}^{t-1}
Q_{\omega_1}^j
\left(
s_m^j,
\hat a_m^j
\right)
\right].
\end{equation}

Accordingly, the loss function for fine-tuning the DT on task $j$ is formulated as
\begin{equation}
\mathcal L^j(\phi,\theta,\eta)
=
\mathcal L^j_{\mathrm{BC}}
+
\lambda_J' \mathcal L^j_{\mathrm{JEPA}}
+
\lambda_Q \mathcal L_Q^j,
\end{equation}
where $\lambda_J'$ and $\lambda_Q$ represent the hyperparameter that balance the loss terms.

\section{Experiments}

Our experiments systematically evaluate the proposed DT-JEPA framework centered on the following three questions:

\begin{itemize}
    \item \textbf{RQ1:} Can DT-JEPA learn transferable knowledge from reward-free datasets to support downstream policy adaptation?

    \item \textbf{RQ2:} Can DT-JEPA effectively transfer the knowledge acquired during the pretraining phase to unseen new tasks?

    \item \textbf{RQ3:} How do the latent future representation $\hat{\mathbf F}_t$ learned by JEPA and online fine-tuning mechanism contribute to the DT-JEPA's performance?
\end{itemize}

\subsection{Baselines}

We compare DT-JEPA with a variety of offline RL and DT-based methods.
\textbf{TD3-BC} \citep{fujimoto2021minimalist}, a widely used offline RL method that combines value-based policy optimization with behavior cloning regularization; \textbf{TD-JEPA} \citep{bagatella2026tdjepa}, which learns
long-term latent-predictive representations from reward-free offline
transitions for zero-shot downstream RL;
\textbf{QT} \citep{pmlr-v235-hu24c}, which augments
Transformer-based conditional sequence modeling with Q-value
regularization;
\textbf{ODT} \citep{zheng2022online}, an online DT that adapts the pretrained sequence policy through limited environment interaction;
\textbf{Prompt DT} \citep{xu2022prompt}, which conditions DT on trajectory demonstrations to support few-shot policy generalization across tasks;
\textbf{VDT} \citep{zheng2025value}, which incorporates value guidance into DT to improve policy optimization;
\textbf{LSDT} \citep{wang2025long}, which jointly models long-term and local trajectory dependencies to enhance sequence-based decision making;
and \textbf{PDT} \citep{xie2023future}, a method that conditions policy learning on true future trajectory information and subsequently adapts the pretrained policy to downstream tasks.

\subsection{Benchmark Datasets and Evaluation Protocol}

We evaluate DT-JEPA on diverse tasks from the D4RL benchmark.
For MuJoCo locomotion tasks, the experiments include four sequential control
environments: \textit{Hopper}, \textit{Walker2d},
\textit{HalfCheetah}, and \textit{Ant}.
For each environment, we consider three offline datasets:
\textit{medium}, \textit{medium-replay}, and \textit{medium-expert},
which differ in trajectory quality and diversity.
Moreover, we evaluate on long-horizon navigation benchmarks
\textit{AntMaze} and \textit{Maze2D}.
For \textit{AntMaze}, we consider four dataset configurations,
including \textit{umaze}, \textit{umaze-diverse},
\textit{medium-play}, and \textit{medium-diverse}.
For \textit{Maze2D}, we evaluate on the \textit{umaze},
\textit{medium}, and \textit{large} environments with progressively challenging navigation. For compact presentation, we use the following abbreviations:
\textit{m/m-r/m-e} for \textit{medium/medium-replay/medium-expert},
\textit{u/u-d/m-p/m-d} for the four AntMaze variants, and
\textit{u/m/l} for \textit{umaze/medium/large} in Maze2D.
Moreover, we use the D4RL normalized return as the primary evaluation metric.
For each environment and dataset, we perform independent training
with multiple random seeds and report the mean normalized return
together with its standard deviation.

\section{Experimental Results}

\noindent\textbf{RQ1:}
{
\setlength{\textfloatsep}{7pt plus 1pt minus 1pt}
\begin{table}[t]
\centering
\caption{
Normalized returns on D4RL locomotion and navigation tasks.
We report the mean performance together with the standard deviation over
5 random seeds.
}
\label{tab:main_results}

\scriptsize
\setlength{\tabcolsep}{3.8pt}
\renewcommand{\arraystretch}{1.07}

\resizebox{\linewidth}{!}{
\begin{tabular}{lccccccccc}
\toprule
\textbf{Dataset}
& \textbf{TD3-BC}
& \textbf{TD-JEPA}
& \textbf{ODT}
& \textbf{QT}
& \textbf{Prompt DT}
& \textbf{VDT}
& \textbf{LSDT}
& \textbf{PDT}
& \textbf{DT-JEPA} \\
\midrule

Hopper-m
& $82.6 \pm 3.8$
& $85.7 \pm 3.4$
& $79.6 \pm 3.5$
& $95.3 \pm 3.6$
& $90.6 \pm 2.9$
& $94.1 \pm 2.5$
& $89.4 \pm 1.9$
& $\mathbf{96.2 \pm 1.8}$
& $\underline{95.4 \pm 1.7}$ \\

Hopper-m-r
& $64.3 \pm 5.6$
& $68.9 \pm 5.0$
& $74.8 \pm 5.1$
& $\underline{97.3 \pm 3.1}$
& $86.1 \pm 3.4$
& $96.1 \pm 3.5$
& $93.5 \pm 2.9$
& $82.9 \pm 3.5$
& $\mathbf{98.3 \pm 2.1}$ \\

Hopper-m-e
& $96.8 \pm 2.7$
& $99.1 \pm 2.4$
& $98.9 \pm 2.5$
& $102.0 \pm 1.8$
& $100.7 \pm 2.1$
& $100.4 \pm 1.7$
& $\underline{104.1 \pm 1.7}$
& $102.4 \pm 1.8$
& $\mathbf{105.2 \pm 1.3}$ \\

\midrule

Walker2d-m
& $78.9 \pm 4.7$
& $81.2 \pm 4.2$
& $74.6 \pm 5.3$
& $82.6 \pm 3.4$
& $79.8 \pm 4.1$
& $81.7 \pm 3.6$
& $\underline{83.4 \pm 3.1}$
& $79.3 \pm 3.9$
& $\mathbf{86.1 \pm 2.8}$ \\

Walker2d-m-r
& $88.3 \pm 3.7$
& $73.6 \pm 5.4$
& $72.5 \pm 4.9$
& $\mathbf{99.0 \pm 3.9}$
& $72.9 \pm 5.6$
& $75.1 \pm 4.8$
& $80.4 \pm 3.9$
& $64.6 \pm 4.3$
& $\underline{94.4 \pm 2.3}$ \\

Walker2d-m-e
& $104.8 \pm 3.0$
& $106.8 \pm 2.6$
& $106.2 \pm 2.7$
& $110.5 \pm 2.0$
& $107.7 \pm 2.4$
& $109.1 \pm 2.1$
& $\underline{111.8 \pm 1.7}$
& $108.6 \pm 2.0$
& $\mathbf{112.3 \pm 1.5}$ \\

\midrule

Halfcheetah-m
& $47.9 \pm 1.2$
& $49.0 \pm 1.1$
& $43.6 \pm 2.4$
& $\underline{51.3 \pm 1.1}$
& $47.1 \pm 1.6$
& $48.5 \pm 1.4$
& ${49.7 \pm 1.1}$
& $40.2 \pm 2.3$
& $\mathbf{52.3 \pm 0.9}$ \\

Halfcheetah-m-r
& $44.1 \pm 1.8$
& $45.5 \pm 1.6$
& $40.8 \pm 2.6$
& $\underline{48.0 \pm 1.4}$
& $43.7 \pm 2.1$
& $41.4 \pm 1.4$
& $47.2 \pm 1.5$
& $32.9 \pm 2.8$
& $\mathbf{49.8 \pm 1.2}$ \\

Halfcheetah-m-e
& $91.4 \pm 1.7$
& $92.8 \pm 1.5$
& $89.7 \pm 2.3$
& $94.0 \pm 1.4$
& $92.1 \pm 1.8$
& $93.5 \pm 1.5$
& $\underline{94.6 \pm 1.3}$
& $93.1 \pm 1.6$
& $\mathbf{96.2 \pm 1.0}$ \\

\midrule

Ant-m
& $87.6 \pm 4.3$
& $90.0 \pm 3.8$
& $82.8 \pm 5.2$
& $91.2 \pm 3.2$
& $87.9 \pm 4.0$
& $90.1 \pm 3.5$
& $\underline{92.3 \pm 3.0}$
& $89.7 \pm 3.6$
& $\mathbf{95.1 \pm 2.4}$ \\

Ant-m-r
& $71.5 \pm 6.3$
& $74.2 \pm 5.7$
& $76.3 \pm 4.7$
& $\underline{78.0 \pm 4.1}$
& $71.9 \pm 5.9$
& $74.7 \pm 5.0$
& $77.1 \pm 4.3$
& $72.6 \pm 5.4$
& $\mathbf{80.6 \pm 3.6}$ \\

Ant-m-e
& $108.7 \pm 3.1$
& $110.2 \pm 2.8$
& $105.9 \pm 3.8$
& $112.1 \pm 2.2$
& $109.4 \pm 2.9$
& $111.2 \pm 2.5$
& $\underline{112.8 \pm 2.1}$
& $110.6 \pm 2.6$
& $\mathbf{114.5 \pm 1.7}$ \\

\midrule

Antmaze-u
& $75.7 \pm 4.9$
& $62.7 \pm 5.1$
& $58.4 \pm 5.6$
& $\mathbf{95.0 \pm 3.6}$
& $65.8 \pm 4.7$
& $\underline{92.1 \pm 4.7}$
& $73.5 \pm 3.7$
& $71.8 \pm 4.0$
& ${77.4 \pm 3.1}$ \\

Antmaze-u-d
& $71.8 \pm 4.6$
& $49.5 \pm 5.8$
& $51.4 \pm 4.5$
& $\underline{94.7 \pm 4.2}$
& $53.6 \pm 5.4$
& $91.3 \pm 4.2$
& $80.3 \pm 3.7$
& $59.4 \pm 4.7$
& $\mathbf{94.9 \pm 2.9}$ \\

Antmaze-m-p
& $9.8 \pm 2.4$
& $35.9 \pm 6.1$
& $32.1 \pm 6.5$
& ${19.8 \pm 4.4}$
& $39.8 \pm 5.8$
& $44.5 \pm 5.1$
& $\mathbf{79.9 \pm 4.8}$
& $46.1 \pm 4.9$
& $\underline{52.9 \pm 4.0}$ \\

Antmaze-m-d
& $3.4 \pm 1.5$
& $27.8 \pm 6.4$
& $24.6 \pm 6.7$
& $\underline{56.2 \pm 4.7}$
& $31.5 \pm 6.0$
& $35.2 \pm 5.5$
& $73.8 \pm 4.1$
& $37.4 \pm 5.1$
& $\mathbf{73.9 \pm 3.1}$ \\

\midrule

Maze2d-u
& $14.3 \pm 4.6$
& $88.9 \pm 4.5$
& $86.2 \pm 4.8$
& $\underline{97.0 \pm 2.9}$
& $91.6 \pm 3.9$
& $89.3 \pm 3.2$
& $9.5 \pm 3.0$
& $94.7 \pm 3.4$
& $\mathbf{99.8 \pm 2.4}$ \\

Maze2d-m
& $60.4 \pm 3.7$
& $81.2 \pm 5.1$
& $78.4 \pm 5.5$
& $\underline{91.5 \pm 3.3}$
& $84.6 \pm 4.6$
& $63.1 \pm 3.8$
& $17.3 \pm 2.5$
& $88.5 \pm 3.8$
& $\mathbf{94.1 \pm 2.9}$ \\

Maze2d-l
& $85.6 \pm 5.9$
& $71.7 \pm 5.8$
& $68.5 \pm 6.1$
& $\underline{84.0 \pm 3.9}$
& $75.4 \pm 5.2$
& $79.0 \pm 4.7$
& $12.6 \pm 3.0$
& $80.4 \pm 4.4$
& $\mathbf{87.2 \pm 3.3}$ \\

\bottomrule
\end{tabular}
}
\end{table}
}
As shown in Table~\ref{tab:main_results}, DT-JEPA achieves the best
performance on 10 out of 12 Gym MuJoCo datasets and remains competitive
across the navigation tasks. In particular, DT-JEPA ranks first on three
of the four \textit{medium-replay} datasets, indicating that predictive
future representations are effective at extracting useful decision
information from heterogeneous offline behaviors. More importantly,
DT-JEPA outperforms PDT on 18 out of 19 datasets, suggesting that
JEPA-based predictive representation learning provides a stronger decision prior for policy learning. Meanwhile, DT-JEPA 
consistently surpasses TD-JEPA, showing the benefit of integrating
latent-predictive representation learning with sequence-based policy
optimization. 
\begin{figure}[h]
\centering
\includegraphics[width=1\textwidth]{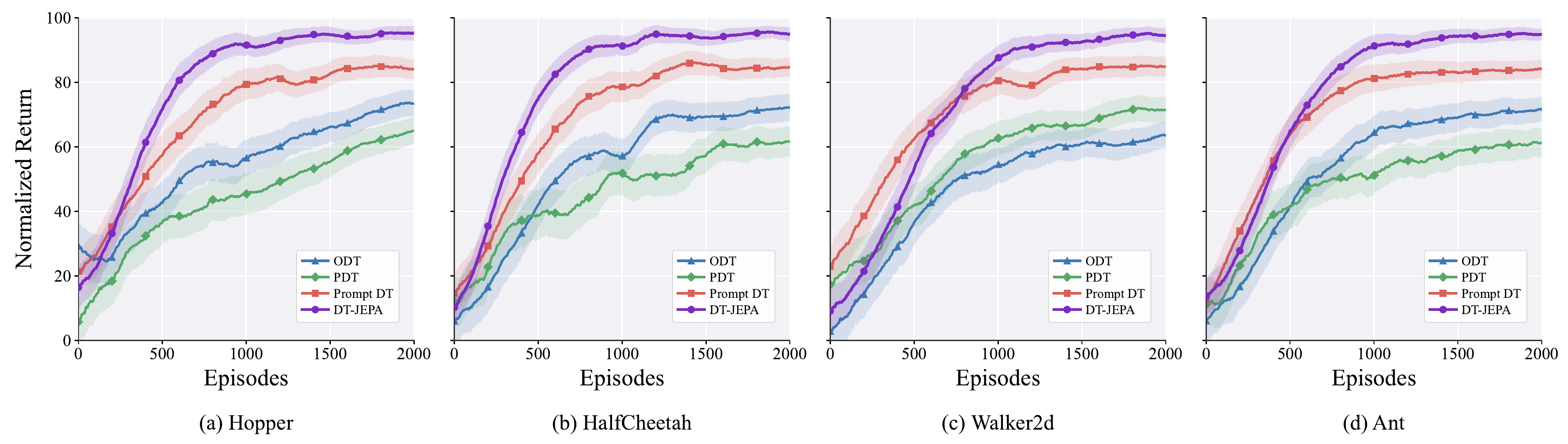}
\caption{
Fine-tuning performance and convergence comparison of DT-JEPA and competing methods on unseen Gym MuJoCO tasks. Performance is evaluated every 10 episodes. We report results averaged over three random seeds, with shaded regions indicating 95$\%$ confidence intervals.
}
\label{fig:finetuning_convergence}
\end{figure}

\begin{figure}[h]
\centering
\includegraphics[width=1\textwidth]{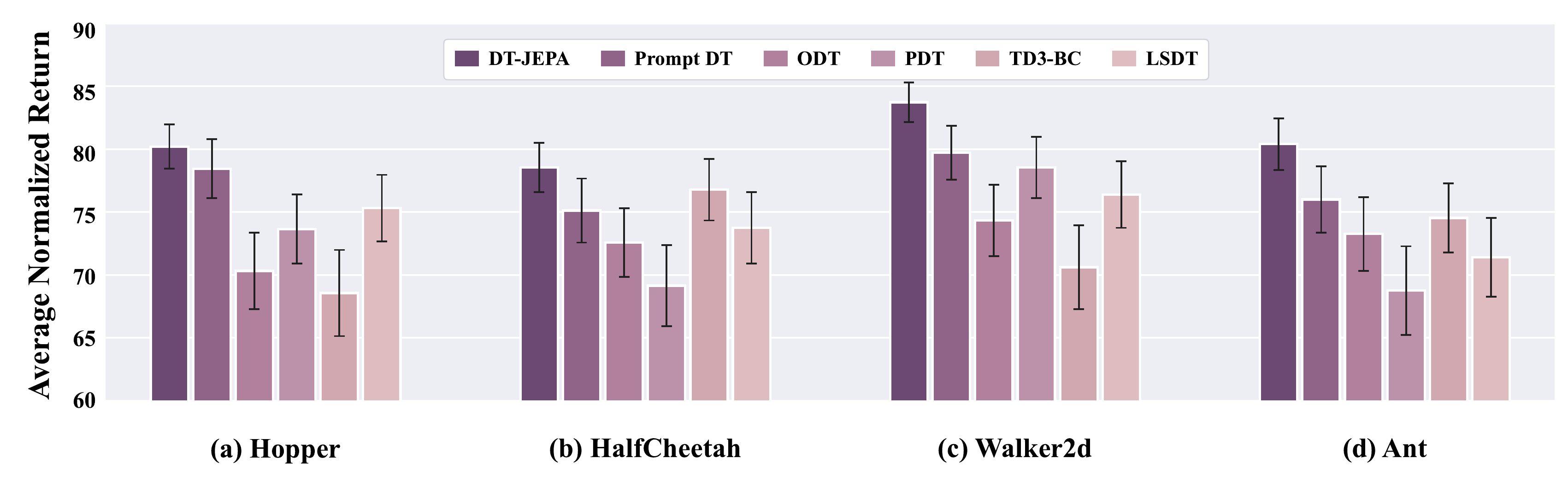}
\caption{
Generalization performance of DT-JEPA and competing methods on unseen downstream tasks. The results report the averaged normalized return achieved on 3 unseen downstream tasks, and the error bars represent 95$\%$ confidence intervals.
}
\label{fig:zero_shot_transfer1}
\end{figure}

\noindent\textbf{RQ2:}
As shown in Figure~\ref{fig:finetuning_convergence}, we first evaluate
the convergence of DT-JEPA on one unseen reward function in each of the
four MuJoCo environments. DT-JEPA reaches high-return regions with fewer
interaction episodes and consistently converges faster than Prompt DT,
ODT, and PDT. This advantage indicates that critic-guided adaptation can
effectively steer the pretrained policy toward new reward objectives,
thereby improving downstream adaptation efficiency.
Figure~\ref{fig:zero_shot_transfer1} further plots the averaged normalized return achieved on 3 unseen reward objectives of the four MuJoCO tasks.  DT-JEPA consistently
outperforms the competing methods across the four MuJoCo tasks,
indicating that the predictive future representations learned during
reward-free pretraining capture reusable dynamics information for the adaptation of diverse downstream tasks.
\begin{table}[t]
\centering
\caption{
Ablation study of DT-JEPA on unseen downstream tasks.
We evaluate the effects on four MuJoCo environments
with unseen reward objectives during pretraining.
Results are reported in terms of normalized return.
}
\label{tab:dtjepa_ablation}

\scriptsize
\resizebox{\linewidth}{!}{
\begin{tabular}{lcccccc}
\toprule
\textbf{Method}
& \textbf{Condition}
& \textbf{Hopper}
& \textbf{Walker2d}
& \textbf{HalfCheetah}
& \textbf{Ant}
& \textbf{Avg.} \\
\midrule

DT
& RTG
& 68.4
& 64.9
& 72.3
& 60.8
& 66.6 \\

DT-RTG-AdaLN
& RTG
& 71.0
& 67.8
& 71.9
& 64.6
& 68.8 \\

DT + Deterministic Future Predictor
& Future
& 76.8
& 74.5
& 80.2
& 71.6
& 75.8 \\

DT-JEPA w/o Critic Guidance
& Future
& 82.1
& 78.6
& 84.8
& 76.0
& 80.4 \\

DT-JEPA w/ Frozen Backbone
& Future
& 83.5
& 80.9
& 84.4
& 76.7
& 81.4 \\

DT-JEPA-Token
& Future
& 84.7
& 80.1
& 87.9
& 79.3
& 83.0 \\

\textbf{DT-JEPA}
& Future
& \textbf{89.4}
& \textbf{85.6}
& \textbf{91.2}
& \textbf{83.1}
& \textbf{87.3} \\

\bottomrule
\end{tabular}
}
\end{table}

\noindent\textbf{RQ3:}
To examine the contribution of each component to downstream
generalization, we conduct ablation studies on unseen reward objectives
across Hopper, Walker2d, HalfCheetah, and Ant. As shown in Table~\ref{tab:dtjepa_ablation}, future-conditioned variants
consistently outperform scalar return-based conditioning, while probabilistic JEPA
further improves over deterministic future prediction. This indicates
that the learned latent future representation $\hat{\mathbf F}_t$
captures informative predictive structure that benefits action generation.
Moreover, AdaLN-based modulation performs better overall than token
injection, suggesting that feature-wise modulation allows the policy to
exploit the conditional information more effectively. Finally, removing critic guidance or freezing the pretrained backbone both degrade performance, showing
that online fine-tuning is essential for adapting the learned predictive
representation and policy to new reward objectives.
\section{Conclusion}

In this paper, we introduced DT-JEPA, a predictive representation learning
framework that integrates JEPA with DT for transferable decision making.
DT-JEPA learns probabilistic future representations from offline trajectories
and incorporates them into action generation through AdaLN. We further
developed a critic-guided fine-tuning mechanism to adapt the pretrained
policy to unseen reward objectives with limited interaction. Theoretically,
we analyze the predictive learning objective and derive a finite-sample
performance bound for future-conditioned policy learning. Experiments show
that DT-JEPA achieves strong performance and efficient adaptation, validating
predictive future representations as an effective condition information for
generalizable sequence decision making.

\subsection*{AI use statement}
Generative AI tools were used solely for language editing and presentation,
including improvements to grammar, clarity, concision, and readability.
They were not used to develop the research methodology, formulate
mathematical claims, implement the proposed method, generate or process
experimental data, conduct experiments, or interpret the results.
All AI-assisted revisions were carefully reviewed and verified by the
authors, who take full responsibility for the final content of the paper.

\bibliography{iclr2027_conference}
\bibliographystyle{iclr2027_conference}

\clearpage
\appendix
\section{Related Work}
\subsection{Decision Transformers and Conditional Sequence Modeling}

Reinforcement learning algorithms face the challenge of long-term credit assignment, which is typically addressed through temporal-difference (TD) learning~\citep{sutton2018reinforcement}. Transformers~\citep{vaswani2017attention}, originally developed for sequence modeling, provide a powerful mechanism for capturing long-range dependencies across trajectories. Decision Transformer (DT)~\citep{chen2021decision} builds on this perspective and formulates reinforcement learning as a conditional sequence modeling problem, representing trajectories as sequences of returns, states, and actions and predicting actions conditioned on desired returns. Prompt-DT~\citep{xu2022prompt} further introduces trajectory demonstrations as prompts to encode task-specific information for few-shot policy generalization. Generalized DT (GDT)~\citep{furuta2022generalized} interprets conditional policy learning through hindsight information matching and extends the conditioning signal beyond returns to future states and other trajectory statistics. Meta-DT~\citep{wang2024metadt} further applies conditional sequence modeling to offline meta-RL by learning task representations through a disentangled world model for few-shot and zero-shot adaptation. These studies suggest that the conditioning information is central to the generalization ability of sequence-based policies.

\subsection{JEPA-based Representation Learning for RL}

Future latent prediction has been widely used in reinforcement learning to learn decision-relevant representations and improve sample efficiency without reconstructing raw observations~\citep{schwarzer2021data,guo2020contrastive}. More recently, JEPA provide a general framework for predictive representation learning by predicting the latent representation of a target from an observed context rather than reconstructing the target in the original input space~\citep{lecun2022path}. This work has been later extended to video representation learning through V-JEPA, where models learn by predicting latent representations of masked spatiotemporal regions~\citep{bardes2024vjepa}. In reinforcement learning, several representative methods have explored latent prediction from different perspectives. Recent methods such as TD-JEPA~\citep{bagatella2026tdjepa} further study reward-free representation learning through policy-conditioned temporal prediction, emphasizing long-term dynamics that can support zero-shot control. Kenneweg et al.~\citep{kenneweg2025jepa} further combine JEPA with PPO, jointly optimizing action-conditioned latent prediction and actor--critic objectives for reinforcement learning from visual observations.

\subsection{Reward-free Pretraining and Zero-shot RL}

Reward-free RL focuses on policy learning from unlabeled datasets without relying on task-specific reward supervision. Forward-Backward (FB) representations learn state-action occupancy representations that support zero-shot policy construction for downstream rewards~\citep{touati2021learning,touati2023zeroshot}. HILP instead learns a temporally structured latent space for discovering and composing goal-directed behaviors from reward-free data~\citep{park2024hilp}. RLDP explores reusable policy representations learned from offline trajectories to facilitate generalization to unseen reward specifications~\citep{jajoo2026rldp}. These approaches highlight different ways of extracting task-agnostic structure from reward-free experience for downstream control. These methods demonstrate that reward-free RL methods can acquire reusable behavioral knowledge without specific reward supervision.

\section{Proofs}

\subsection{Proof of Theorem 1}

According to the chain rule of mutual information, we have
\begin{align}
I(\mathbf H_t;\tau_{\geq t},\mathbf F_t)
&=
I(\mathbf H_t;\tau_{\geq t})
+
I(\mathbf H_t;\mathbf F_t\mid\tau_{\geq t})
\nonumber\\
&=
I(\mathbf H_t;\mathbf F_t)
+
I(\mathbf H_t;\tau_{\geq t}\mid\mathbf F_t).
\label{eq:mi_chain_rule}
\end{align}

Since $\mathbf F_t$ is obtained from $\tau_{\geq t}$ through the
target encoder, once $\tau_{\geq t}$ is given,
$\mathbf F_t$ contains no additional information about $\mathbf H_t$.
Therefore,
\begin{equation}
I(\mathbf H_t;\mathbf F_t\mid\tau_{\geq t})=0.
\label{eq:future_independence}
\end{equation}

Substituting ~\eqref{eq:future_independence} into
~\eqref{eq:mi_chain_rule} yields
\begin{equation}
I(\mathbf H_t;\tau_{\geq t})
=
I(\mathbf H_t;\mathbf F_t)
+
I(\mathbf H_t;\tau_{\geq t}\mid\mathbf F_t).
\label{eq:mi_decomposition}
\end{equation}

Since mutual information is non-negative,
\begin{equation}
I(\mathbf H_t;\tau_{\geq t}\mid\mathbf F_t)
\geq 0,
\end{equation}
we obtain
\begin{equation}
I(\mathbf H_t;\tau_{\geq t})
\geq
I(\mathbf F_t;\mathbf H_t).
\label{eq:future_mi_upper_bound}
\end{equation}

According to the definition of mutual information, the mutual
information between $\mathbf F_t$ and $\mathbf H_t$ is
\begin{equation}
I(\mathbf F_t;\mathbf H_t)
=
\mathbb E_{p(\mathbf F_t,\mathbf H_t)}
\left[
\log
\frac{
p(\mathbf F_t,\mathbf H_t)
}{
p(\mathbf F_t)p(\mathbf H_t)
}
\right].
\label{eq:mi_def_joint}
\end{equation}

The joint distribution can be factorized as
\begin{equation}
p(\mathbf F_t,\mathbf H_t)
=
p(\mathbf F_t\mid\mathbf H_t)
p(\mathbf H_t).
\label{eq:joint_factorization}
\end{equation}

Substituting ~\eqref{eq:joint_factorization} into
~\eqref{eq:mi_def_joint}, we obtain
\begin{align}
I(\mathbf F_t;\mathbf H_t)
&=
\mathbb E_{p(\mathbf F_t,\mathbf H_t)}
\left[
\log
\frac{
p(\mathbf F_t\mid\mathbf H_t)
p(\mathbf H_t)
}{
p(\mathbf F_t)
p(\mathbf H_t)
}
\right]
\nonumber\\
&=
\mathbb E_{p(\mathbf F_t,\mathbf H_t)}
\left[
\log
\frac{
p(\mathbf F_t\mid\mathbf H_t)
}{
p(\mathbf F_t)
}
\right].
\label{eq:mi_ratio}
\end{align}

Further expanding the logarithmic term gives
\begin{align}
I(\mathbf F_t;\mathbf H_t)
&=
\mathbb E_{p(\mathbf F_t,\mathbf H_t)}
\left[
\log p(\mathbf F_t\mid\mathbf H_t)
-
\log p(\mathbf F_t)
\right]
\nonumber\\
&=
\mathbb E_{p(\mathbf F_t,\mathbf H_t)}
\left[
\log p(\mathbf F_t\mid\mathbf H_t)
\right]-
\mathbb E_{p(\mathbf F_t,\mathbf H_t)}
\left[
\log p(\mathbf F_t)
\right].
\label{eq:mi_expand_log}
\end{align}

Since $\log p(\mathbf F_t)$ is independent of $\mathbf H_t$,
marginalizing over $\mathbf H_t$ gives
\begin{align}
&
\mathbb E_{p(\mathbf F_t,\mathbf H_t)}
\left[
\log p(\mathbf F_t)
\right]
\nonumber\\
&=
\int
p(\mathbf F_t,\mathbf H_t)
\log p(\mathbf F_t)
\,d\mathbf F_t\,d\mathbf H_t
\nonumber\\
&=
\int
\left[
\int
p(\mathbf F_t,\mathbf H_t)
\,d\mathbf H_t
\right]
\log p(\mathbf F_t)
\,d\mathbf F_t
\nonumber\\
&=
\int
p(\mathbf F_t)
\log p(\mathbf F_t)
\,d\mathbf F_t
\nonumber\\
&=
-H(\mathbf F_t).
\label{eq:marginal_entropy_conversion}
\end{align}

Therefore, ~\eqref{eq:mi_expand_log} can be rewritten as
\begin{equation}
I(\mathbf F_t;\mathbf H_t)
=
H(\mathbf F_t)
+
\mathbb E_{p(\mathbf F_t,\mathbf H_t)}
\left[
\log p(\mathbf F_t\mid\mathbf H_t)
\right].
\label{eq:mi_entropy_form}
\end{equation}

Equivalently,
\begin{equation}
H(\mathbf F_t\mid\mathbf H_t)
=
-
\mathbb E_{p(\mathbf F_t,\mathbf H_t)}
\left[
\log p(\mathbf F_t\mid\mathbf H_t)
\right],
\end{equation}
and thus
\begin{equation}
I(\mathbf F_t;\mathbf H_t)
=
H(\mathbf F_t)
-
H(\mathbf F_t\mid\mathbf H_t).
\label{eq:mi_entropy_definition}
\end{equation}

However, the true distribution
$p(\mathbf F_t\mid\mathbf H_t)$
is generally difficult to obtain directly.
Therefore, we introduce a predictive distribution
$q_{\eta}(\mathbf F_t\mid\mathbf H_t)$
parameterized by $\eta$ to approximate it.

For any given $\mathbf H_t$, according to the non-negativity
of the Kullback--Leibler divergence, we have
\begin{equation}
D_{\mathrm{KL}}
\left(
p(\mathbf F_t\mid\mathbf H_t)
\parallel
q_{\eta}(\mathbf F_t\mid\mathbf H_t)
\right)
\geq 0.
\label{eq:kl_nonnegative}
\end{equation}

Expanding the KL divergence gives
\begin{align}
&D_{\mathrm{KL}}
\left(
p(\mathbf F_t\mid\mathbf H_t)
\parallel
q_{\eta}(\mathbf F_t\mid\mathbf H_t)
\right)
\nonumber\\
&=
\int
p(\mathbf F_t\mid\mathbf H_t)
\log
\frac{
p(\mathbf F_t\mid\mathbf H_t)
}{
q_{\eta}(\mathbf F_t\mid\mathbf H_t)
}
\,d\mathbf F_t
\nonumber\\
&=
\mathbb E_{p(\mathbf F_t\mid\mathbf H_t)}
\left[
\log p(\mathbf F_t\mid\mathbf H_t)
\right]-
\mathbb E_{p(\mathbf F_t\mid\mathbf H_t)}
\left[
\log q_{\eta}(\mathbf F_t\mid\mathbf H_t)
\right]
\geq 0.
\label{eq:kl_expand}
\end{align}

It follows that
\begin{align}
&
\mathbb E_{p(\mathbf F_t\mid\mathbf H_t)}
\left[
\log p(\mathbf F_t\mid\mathbf H_t)
\right]
\geq
\mathbb E_{p(\mathbf F_t\mid\mathbf H_t)}
\left[
\log q_{\eta}(\mathbf F_t\mid\mathbf H_t)
\right].
\label{eq:variational_inequality}
\end{align}

Further taking the expectation over $p(\mathbf H_t)$ gives
\begin{align}
&
\mathbb E_{p(\mathbf H_t)}
\mathbb E_{p(\mathbf F_t\mid\mathbf H_t)}
\left[
\log p(\mathbf F_t\mid\mathbf H_t)
\right]
\geq
\mathbb E_{p(\mathbf H_t)}
\mathbb E_{p(\mathbf F_t\mid\mathbf H_t)}
\left[
\log q_{\eta}(\mathbf F_t\mid\mathbf H_t)
\right].
\label{eq:expect_over_history}
\end{align}

Since
\begin{equation}
p(\mathbf F_t,\mathbf H_t)
=
p(\mathbf H_t)
p(\mathbf F_t\mid\mathbf H_t),
\end{equation}
~\eqref{eq:expect_over_history} is equivalent to
\begin{align}
&
\mathbb E_{p(\mathbf F_t,\mathbf H_t)}
\left[
\log p(\mathbf F_t\mid\mathbf H_t)
\right]
\geq
\mathbb E_{p(\mathbf F_t,\mathbf H_t)}
\left[
\log q_{\eta}(\mathbf F_t\mid\mathbf H_t)
\right].
\label{eq:joint_variational_inequality}
\end{align}

Substituting ~\eqref{eq:joint_variational_inequality}
into ~\eqref{eq:mi_entropy_form}, we obtain
\begin{align}
I(\mathbf F_t;\mathbf H_t)
&=
H(\mathbf F_t)
+
\mathbb E_{p(\mathbf F_t,\mathbf H_t)}
\left[
\log p(\mathbf F_t\mid\mathbf H_t)
\right]
\nonumber\\
&\geq
H(\mathbf F_t)
+
\mathbb E_{p(\mathbf F_t,\mathbf H_t)}
\left[
\log q_{\eta}(\mathbf F_t\mid\mathbf H_t)
\right].
\label{eq:final_mi_bound}
\end{align}

Combining ~\eqref{eq:future_mi_upper_bound}
and ~\eqref{eq:final_mi_bound}, we finally obtain
\begin{align}
I(\mathbf H_t;\tau_{\geq t})
&\geq
I(\mathbf F_t;\mathbf H_t)
\nonumber\\
&\geq
\mathbb E_{p(\mathbf F_t,\mathbf H_t)}
\left[
\log q_{\eta}(\mathbf F_t\mid\mathbf H_t)
\right]
+
H(\mathbf F_t).
\label{eq:final_mi_lower_bound}
\end{align}

\subsection{Proof of Theorem~2}
\label{app:proof_predictive_conditioning}

\textbf{Theorem 2.}
\label{thm:pfc_sample_complexity}
\textit{
Consider a finite-horizon MDP with horizon $H$ and rewards in $[0,1]$.
For a given predictor $q_\eta(F\mid x)$, let
$\bar{\pi}_\eta^{\rm P}$ and $\bar{\pi}_{\eta,\hat{\phi}}$
denote the population and learned marginal predictive-conditioned
policies~\citep{brandfonbrener2022does}. Under bounded occupancy mismatch $C_F$, finite policy class
$\Pi_F$, bounded log-likelihood variation $c_F$, and approximation
error $\epsilon_{\rm app}^{F}$, with probability at least $1-\delta$,
}
\begin{equation}
J(\bar{\pi}_\eta^{\rm P})
-
J(\bar{\pi}_{\eta,\hat{\phi}})
\leq
\mathcal O\!\left(
C_F H^2
\left[
\sqrt{c_F}
\left(
\frac{\log(|\Pi_F|/\delta)}{N}
\right)^{1/4}
+
\sqrt{\epsilon_{\rm app}^{F}}
\right]
\right).
\end{equation}

\textit{Proof.}
Let $\beta$ denote the behavior policy
that induces the offline trajectory distribution $\mathcal D$.
We condition throughout the proof on a given predictive model
$q_\eta(\hat F_t\mid H_t)$, with
$H_t=f_\theta(\tau_{t-K+1:t}),
\hat F_t\sim q_\eta(\cdot\mid H_t)$.
For convenience, we use
$\pi_\phi(a_t\mid\tau_{t-K+1:t},\hat F_t)$
to denote the conditional action distribution induced by the
DT-JEPA policy network $\phi$. For continuous-control tasks, this
distribution can be interpreted as a fixed-covariance Gaussian, i.e.,
\begin{align}
\pi_\phi
\left(
a_t\mid\tau_{t-K+1:t},\hat F_t
\right)
&=
\mathcal N
\left(
a_t;
\hat a_t,
\sigma_a^2 I
\right).
\end{align}
Thus, the negative log-likelihood satisfies
\begin{align}
-\log
\pi_\phi
\left(
a_t\mid\tau_{t-K+1:t},\hat F_t
\right)
&=
\frac{1}{2\sigma_a^2}
\left\|
a_t-\hat a_t
\right\|_2^2+C,
\label{eq:pfc_gaussian_bc}
\end{align}
where $C$ does not depend on $\phi$. Hence,
minimizing the negative log-likelihood is equivalent to minimizing the
behavior-cloning objective in Eq.~(13). Let
$d_\beta(\tau_{t-K+1:t})$ denote the average occupancy distribution
of historical trajectory segments generated by $\beta$. Since
$\hat F_t$ is sampled from the predictor using $H_t$, the joint
training distribution can be written as
\begin{align}
P_\eta
\left(
\tau_{t-K+1:t},
a_t,
\hat F_t
\right)
=
d_\beta(\tau_{t-K+1:t})\,
\beta(a_t\mid\tau_{t-K+1:t})\,
q_\eta(\hat F_t\mid H_t).
\label{eq:pfc_training_distribution}
\end{align}
Consequently, conditioned on the observable historical trajectory
and the predicted future representation, we have
\begin{align}
P_\eta
\left(
a_t
\mid
\tau_{t-K+1:t},
\hat F_t
\right)
=
\frac{
d_\beta(\tau_{t-K+1:t})
\beta(a_t\mid\tau_{t-K+1:t})
q_\eta(\hat F_t\mid H_t)
}{
d_\beta(\tau_{t-K+1:t})
q_\eta(\hat F_t\mid H_t)
}
=
\beta(a_t\mid\tau_{t-K+1:t}).
\label{eq:pfc_conditional_action}
\end{align}

For any predictive future-conditioned policy
$\pi_\phi\in\Pi_F$, define its population negative log-likelihood as
\begin{align}
\mathcal R_\eta(\pi_\phi)
&=
-
\mathbb E_{\substack{
\tau_{t-K+1:t}\sim d_\beta,\\
\hat F_t\sim q_\eta(\cdot\mid H_t),\\
a_t\sim\beta(\cdot\mid\tau_{t-K+1:t})
}}
\left[
\log
\pi_\phi
\left(
a_t\mid\tau_{t-K+1:t},\hat F_t
\right)
\right].
\label{eq:pfc_population_risk}
\end{align}
For each
$\tau_{t-K+1:t}$ and $\hat F_t$, the corresponding cross entropy can be
decomposed as
\begin{align}
&-
\mathbb E_{
a_t\sim\beta(\cdot\mid\tau_{t-K+1:t})
}
\left[
\log
\pi_\phi
\left(
a_t\mid\tau_{t-K+1:t},\hat F_t
\right)
\right]
\nonumber\\
&=
-
\mathbb E_{
a_t\sim\beta(\cdot\mid\tau_{t-K+1:t})
}
\left[
\log
\beta
\left(
a_t\mid\tau_{t-K+1:t}
\right)
\right]
\nonumber\\
&\quad+
\mathbb E_{
a_t\sim\beta(\cdot\mid\tau_{t-K+1:t})
}
\left[
\log
\frac{
\beta(a_t\mid\tau_{t-K+1:t})
}{
\pi_\phi(a_t\mid\tau_{t-K+1:t},\hat F_t)
}
\right]
\nonumber\\
&=
H\!\left(
\beta(\cdot\mid\tau_{t-K+1:t})
\right)
+
D_{\rm KL}
\left(
\beta(\cdot\mid\tau_{t-K+1:t})
\Vert
\pi_\phi(\cdot\mid\tau_{t-K+1:t},\hat F_t)
\right).
\label{eq:pfc_cross_entropy_decomposition}
\end{align}
Taking expectation over the historical trajectory and the predicted
future representation gives
\begin{align}
\mathcal R_\eta(\pi_\phi)
&=
\mathcal H_\beta
+
\mathcal E_\eta(\pi_\phi),
\label{eq:pfc_error_definition}
\end{align}
where \begin{align}\mathcal H_\beta
=
\mathbb E_{\tau_{t-K+1:t}\sim d_\beta}
\left[
H\!\left(
\beta(\cdot\mid\tau_{t-K+1:t})
\right)
\right],
\end{align}
\begin{align}
\mathcal E_\eta(\pi_\phi)
=
\mathbb E_{\substack{
\tau_{t-K+1:t}\sim d_\beta,
\hat F_t\sim q_\eta(\cdot\mid H_t)
}}
\left[
D_{\rm KL}
\left(
\beta(\cdot\mid\tau_{t-K+1:t})
\Vert
\pi_\phi(\cdot\mid\tau_{t-K+1:t},\hat F_t)
\right)
\right].
\end{align} 

Since $\mathcal H_\beta$ is independent of the policy parameters,
the population predictive-conditioned policy satisfies
\begin{align}
\pi_\eta^{\rm P}
&\in
\arg\min_{\pi_\phi\in\Pi_F}
\mathcal R_\eta(\pi_\phi)
=
\arg\min_{\pi_\phi\in\Pi_F}
\mathcal E_\eta(\pi_\phi),
\nonumber\\
\mathcal E_\eta(\pi_\eta^{\rm P})
&=
\inf_{\pi_\phi\in\Pi_F}
\mathcal E_\eta(\pi_\phi)
\leq
\epsilon_{\rm app}^{F}.
\label{eq:pfc_population_policy}
\end{align}

Since the predicted future representations are sampled from the
same predictor during policy learning and policy execution, we have
\begin{align}
\bar{\pi}_\eta^{\rm P}
(a_t\mid\tau_{t-K+1:t})
&=
\int
q_\eta(\hat F_t\mid H_t)
\pi_\eta^{\rm P}
(a_t\mid\tau_{t-K+1:t},\hat F_t)
\,d\hat F_t,
\nonumber\\
\bar{\pi}_{\eta,\hat\phi}
(a_t\mid\tau_{t-K+1:t})
&=
\int
q_\eta(\hat F_t\mid H_t)
\pi_{\hat\phi}
(a_t\mid\tau_{t-K+1:t},\hat F_t)
\,d\hat F_t.
\label{eq:pfc_marginal_policies}
\end{align}

Next, we relate the discrepancy between these two policies.
For timestep $h$, let
$d_{\bar{\pi}_\eta^{\rm P}}^h
(\tau_{h-K+1:h})
=
P_{\bar{\pi}_\eta^{\rm P}}
\left(
\tau_{h-K+1:h}
\right),$
$d_{\bar{\pi}_{\eta,\hat\phi}}^h
(\tau_{h-K+1:h})
=
P_{\bar{\pi}_{\eta,\hat\phi}}
\left(
\tau_{h-K+1:h}
\right).$
We define
$\Delta_h
=
\left\|
d_{\bar{\pi}_\eta^{\rm P}}^h
-
d_{\bar{\pi}_{\eta,\hat\phi}}^h
\right\|_1$,
$\delta_h
=
2\,
\mathbb E_{
\tau_{h-K+1:h}
\sim
d_{\bar{\pi}_\eta^{\rm P}}^h
}
\left[
D_{\rm TV}
\left(
\bar{\pi}_\eta^{\rm P}
(\cdot\mid\tau_{h-K+1:h}),
\bar{\pi}_{\eta,\hat\phi}
(\cdot\mid\tau_{h-K+1:h})
\right)
\right]$.
Since both policies start from the same initial-state distribution,
$\Delta_1=0$.
Let
$P_K(\tau_{h-K+2:h+1}\mid\tau_{h-K+1:h},a_h)$
denote the transition kernel over the historical trajectory segment
induced by the environment transition
$P(s_{h+1}\mid s_h,a_h)$.
For $h>1$, the corresponding occupancy distributions satisfy
\begin{align}
&
d_{\bar{\pi}_\eta^{\rm P}}^h
(\tau_{h-K+1:h})
\nonumber\\
&=
\int
d_{\bar{\pi}_\eta^{\rm P}}^{h-1}
(\tau_{h-K:h-1})
\int
\bar{\pi}_\eta^{\rm P}
(a_{h-1}\mid\tau_{h-K:h-1})
\nonumber\\
&\qquad\qquad\times
P_K
(
\tau_{h-K+1:h}
\mid
\tau_{h-K:h-1},
a_{h-1}
)
\,da_{h-1}\,
d\tau_{h-K:h-1},
\nonumber\\
&
d_{\bar{\pi}_{\eta,\hat\phi}}^h
(\tau_{h-K+1:h})
\nonumber\\
&=
\int
d_{\bar{\pi}_{\eta,\hat\phi}}^{h-1}
(\tau_{h-K:h-1})
\int
\bar{\pi}_{\eta,\hat\phi}
(a_{h-1}\mid\tau_{h-K:h-1})
\nonumber\\
&\qquad\qquad\times
P_K
(
\tau_{h-K+1:h}
\mid
\tau_{h-K:h-1},
a_{h-1}
)
\,da_{h-1}\,
d\tau_{h-K:h-1}.
\label{eq:pfc_history_recursion}
\end{align}

Expanding the definitions and adding and subtracting the cross term
$d_{\bar{\pi}_\eta^{\rm P}}^{h-1}
(\tau_{h-K:h-1})
\bar{\pi}_{\eta,\hat\phi}
(a_{h-1}\mid\tau_{h-K:h-1})$,
we obtain
\begin{align}
\Delta_h
&=
\int
\left|
d_{\bar{\pi}_\eta^{\rm P}}^h
(\tau_{h-K+1:h})
-
d_{\bar{\pi}_{\eta,\hat\phi}}^h
(\tau_{h-K+1:h})
\right|
d\tau_{h-K+1:h}
\nonumber\\
&\leq
\int
\Bigg|
\int
d_{\bar{\pi}_\eta^{\rm P}}^{h-1}
(\tau_{h-K:h-1})
\int
\Big[
\bar{\pi}_\eta^{\rm P}
(a_{h-1}\mid\tau_{h-K:h-1})
\nonumber\\
&\qquad\qquad-
\bar{\pi}_{\eta,\hat\phi}
(a_{h-1}\mid\tau_{h-K:h-1})
\Big]
P_K
(
\tau_{h-K+1:h}
\mid
\tau_{h-K:h-1},
a_{h-1}
)
\,da_{h-1}\,
d\tau_{h-K:h-1}
\Bigg|
d\tau_{h-K+1:h}
\nonumber\\
&\quad+
\int
\Bigg|
\int
\Big[
d_{\bar{\pi}_\eta^{\rm P}}^{h-1}
(\tau_{h-K:h-1})
-
d_{\bar{\pi}_{\eta,\hat\phi}}^{h-1}
(\tau_{h-K:h-1})
\Big]
\nonumber\\
&\qquad\qquad\times
\int
\bar{\pi}_{\eta,\hat\phi}
(a_{h-1}\mid\tau_{h-K:h-1})
P_K
(
\tau_{h-K+1:h}
\mid
\tau_{h-K:h-1},
a_{h-1}
)
\,da_{h-1}\,
d\tau_{h-K:h-1}
\Bigg|
d\tau_{h-K+1:h}
\nonumber\\
&\leq
\int
d_{\bar{\pi}_\eta^{\rm P}}^{h-1}
(\tau_{h-K:h-1})
\int
\left|
\bar{\pi}_\eta^{\rm P}
(a_{h-1}\mid\tau_{h-K:h-1})
-
\bar{\pi}_{\eta,\hat\phi}
(a_{h-1}\mid\tau_{h-K:h-1})
\right|
\nonumber\\
&\qquad\qquad\qquad\qquad
da_{h-1}\,
d\tau_{h-K:h-1}
+
\Delta_{h-1}
\nonumber\\
&=
2\,
\mathbb E_{
\tau_{h-K:h-1}
\sim
d_{\bar{\pi}_\eta^{\rm P}}^{h-1}
}
\left[
D_{\rm TV}
\left(
\bar{\pi}_\eta^{\rm P}
(\cdot\mid\tau_{h-K:h-1}),
\bar{\pi}_{\eta,\hat\phi}
(\cdot\mid\tau_{h-K:h-1})
\right)
\right]
+
\Delta_{h-1}
\nonumber\\
&=
\delta_{h-1}+\Delta_{h-1}.
\label{eq:pfc_occupancy_recursion}
\end{align}
Repeatedly applying the recursion and using $\Delta_1=0$, we have
\begin{align}
&\Delta_h
\leq
\delta_{h-1}+\delta_{h-2}+\cdots+\delta_1
=
\sum_{j=1}^{h-1}\delta_j.
\label{eq:pfc_occupancy_sum}
\end{align}

Let
$d_{\bar{\pi}_\eta^{\rm P}}
(\tau_{t-K+1:t})$
and
$d_{\bar{\pi}_{\eta,\hat\phi}}
(\tau_{t-K+1:t})$
denote the corresponding average occupancy distributions over the
horizon. By the definition of $\delta_h$, we therefore have
\begin{align}
&
\left\|
d_{\bar{\pi}_\eta^{\rm P}}
-
d_{\bar{\pi}_{\eta,\hat\phi}}
\right\|_1\leq
\frac{1}{T}
\sum_{h=1}^{T}\Delta_h
\nonumber\\
&\leq
2T\,
\mathbb E_{
\tau_{t-K+1:t}
\sim
d_{\bar{\pi}_\eta^{\rm P}}
}
\left[
D_{\rm TV}
\left(
\bar{\pi}_\eta^{\rm P}
(\cdot\mid\tau_{t-K+1:t}),
\bar{\pi}_{\eta,\hat\phi}
(\cdot\mid\tau_{t-K+1:t})
\right)
\right].
\label{eq:pfc_occupancy_bound}
\end{align}

We now relate the occupancy discrepancy to the return difference.
Since rewards are bounded, the difference between the
expected rewards at timestep $h$ is bounded by the discrepancy
between the corresponding history-action distributions. Thus,
using the above recursion,
\begin{align}
&
J(\bar{\pi}_\eta^{\rm P})
-
J(\bar{\pi}_{\eta,\hat\phi})
\nonumber\\
&\leq
\sum_{h=1}^{T}
\left(
\Delta_h+\delta_h
\right)
\leq
\sum_{h=1}^{T}\delta_h
+
\sum_{h=1}^{T}
\sum_{j=1}^{h-1}\delta_j
\leq
T
\sum_{h=1}^{T}\delta_h
\nonumber\\
&=
2T^2
\mathbb E_{
\tau_{t-K+1:t}
\sim
d_{\bar{\pi}_\eta^{\rm P}}
}
\left[
D_{\rm TV}
\left(
\bar{\pi}_\eta^{\rm P}
(\cdot\mid\tau_{t-K+1:t}),
\bar{\pi}_{\eta,\hat\phi}
(\cdot\mid\tau_{t-K+1:t})
\right)
\right].
\label{eq:pfc_return_tv_population}
\end{align}

To characterize the effect of distribution shift between the
population policy and the offline behavior data, define
$C_F
:=
\operatorname*{ess\,sup}_{\tau_{t-K+1:t}}
d_{\bar{\pi}_\eta^{\rm P}}(\tau_{t-K+1:t})
/
d_\beta(\tau_{t-K+1:t})$,
whenever
$d_{\bar{\pi}_\eta^{\rm P}}\ll d_\beta$.
When the occupancy mismatch coefficient $C_F$ is finite, the
population occupancy can be related to the offline behavior
distribution as,
\begin{align}
&
J(\bar{\pi}_\eta^{\rm P})
-
J(\bar{\pi}_{\eta,\hat\phi})
\nonumber\\
&\leq
2C_FT^2
\mathbb E_{
\tau_{t-K+1:t}
\sim d_\beta
}
\left[
D_{\rm TV}
\left(
\bar{\pi}_\eta^{\rm P}
(\cdot\mid\tau_{t-K+1:t}),
\bar{\pi}_{\eta,\hat\phi}
(\cdot\mid\tau_{t-K+1:t})
\right)
\right].
\label{eq:pfc_return_tv_behavior}
\end{align}

We next expand the total variation discrepancy between the two
marginal predictive-conditioned policies. From
~\eqref{eq:pfc_marginal_policies},
\begin{align}
&2\,
\mathbb E_{
\tau_{t-K+1:t}\sim d_\beta
}
\left[
D_{\rm TV}
\left(
\bar{\pi}_\eta^{\rm P}
(\cdot\mid\tau_{t-K+1:t}),
\bar{\pi}_{\eta,\hat\phi}
(\cdot\mid\tau_{t-K+1:t})
\right)
\right]
\nonumber\\
&=
\mathbb E_{
\tau_{t-K+1:t}\sim d_\beta
}
\Bigg[
\int
\Bigg|
\bar{\pi}_\eta^{\rm P}
(a_t\mid\tau_{t-K+1:t})
-
\bar{\pi}_{\eta,\hat\phi}
(a_t\mid\tau_{t-K+1:t})
\Bigg|
\,da_t
\Bigg]
\nonumber\\
&=
\mathbb E_{
\tau_{t-K+1:t}\sim d_\beta
}
\Bigg[
\int
\Bigg|
\int
q_\eta(\hat F_t\mid H_t)
\Big[
\pi_\eta^{\rm P}
(a_t\mid\tau_{t-K+1:t},\hat F_t)
-
\pi_{\hat\phi}
(a_t\mid\tau_{t-K+1:t},\hat F_t)
\Big]
\,d\hat F_t
\Bigg|
\,da_t
\Bigg]
\nonumber\\
&\leq
\mathbb E_{
\tau_{t-K+1:t}\sim d_\beta
}
\Bigg[
\int
q_\eta(\hat F_t\mid H_t)
\int
\left|
\pi_\eta^{\rm P}
(a_t\mid\tau_{t-K+1:t},\hat F_t)
-
\pi_{\hat\phi}
(a_t\mid\tau_{t-K+1:t},\hat F_t)
\right|
\,da_t\,d\hat F_t
\Bigg]
\nonumber\\
&=
2\,
\mathbb E_{\substack{
\tau_{t-K+1:t}\sim d_\beta,\\
\hat F_t\sim q_\eta(\cdot\mid H_t)
}}
\left[
D_{\rm TV}
\left(
\pi_\eta^{\rm P}
(\cdot\mid\tau_{t-K+1:t},\hat F_t),
\pi_{\hat\phi}
(\cdot\mid\tau_{t-K+1:t},\hat F_t)
\right)
\right].
\label{eq:pfc_mixture_tv}
\end{align}

Using the behavior policy as an intermediate distribution, the
triangle inequality gives
\begin{align}
&
D_{\rm TV}
\left(
\pi_\eta^{\rm P}
(\cdot\mid\tau_{t-K+1:t},\hat F_t),
\pi_{\hat\phi}
(\cdot\mid\tau_{t-K+1:t},\hat F_t)
\right)
\nonumber\\
&\leq
D_{\rm TV}
\left(
\pi_\eta^{\rm P}
(\cdot\mid\tau_{t-K+1:t},\hat F_t),
\beta(\cdot\mid\tau_{t-K+1:t})
\right)
\nonumber\\
&\quad+
D_{\rm TV}
\left(
\beta(\cdot\mid\tau_{t-K+1:t}),
\pi_{\hat\phi}
(\cdot\mid\tau_{t-K+1:t},\hat F_t)
\right).
\label{eq:pfc_tv_triangle}
\end{align}
Applying Pinsker's inequality,
$D_{\rm TV}(P,Q)
\leq
\sqrt{\frac{1}{2}D_{\rm KL}(P\Vert Q)}$,
followed by Jensen's inequality, yields
\begin{align}
&2\,
\mathbb E_{
\tau_{t-K+1:t}\sim d_\beta
}
\left[
D_{\rm TV}
\left(
\bar{\pi}_\eta^{\rm P}
(\cdot\mid\tau_{t-K+1:t}),
\bar{\pi}_{\eta,\hat\phi}
(\cdot\mid\tau_{t-K+1:t})
\right)
\right]
\nonumber\\
&\leq
2\,
\mathbb E_{\tau_{t-K+1:t},\hat F_t}
\left[
\sqrt{
\frac{1}{2}
D_{\rm KL}
\left(
\beta(\cdot\mid\tau_{t-K+1:t})
\Vert
\pi_\eta^{\rm P}
(\cdot\mid\tau_{t-K+1:t},\hat F_t)
\right)
}
\right]
\nonumber\\
&\quad+
2\,
\mathbb E_{\tau_{t-K+1:t},\hat F_t}
\left[
\sqrt{
\frac{1}{2}
D_{\rm KL}
\left(
\beta(\cdot\mid\tau_{t-K+1:t})
\Vert
\pi_{\hat\phi}
(\cdot\mid\tau_{t-K+1:t},\hat F_t)
\right)
}
\right]
\nonumber\\
&\leq
\sqrt{
2\,
\mathcal E_\eta(\pi_\eta^{\rm P})
}
+
\sqrt{
2\,
\mathcal E_\eta(\pi_{\hat\phi})
}
\nonumber\\
&\leq
\sqrt{
2\epsilon_{\rm app}^{F}
}
+
\sqrt{
2\mathcal E_\eta(\pi_{\hat\phi})
}.
\label{eq:pfc_tv_to_kl}
\end{align}

Combining ~\eqref{eq:pfc_return_tv_behavior} and
\eqref{eq:pfc_tv_to_kl}, we finally have
\begin{align}
&
J(\bar{\pi}_\eta^{\rm P})
-
J(\bar{\pi}_{\eta,\hat\phi})
\nonumber\\
&\leq
C_FT^2
\left[
\sqrt{
2\epsilon_{\rm app}^{F}
}
+
\sqrt{
2\mathcal E_\eta(\pi_{\hat\phi})
}
\right].
\label{eq:pfc_performance_reduction}
\end{align}

Thus,
given $N$ independently sampled trajectory segments from the behavior
trajectory distribution,
the empirical predictive-conditioned risk is
\begin{align}
\widehat{\mathcal R}_\eta(\pi_\phi)
&=
-
\frac{1}{N}
\sum_{i=1}^{N}
\log
\pi_\phi
\left(
a_{t_i}
\mid
\tau_{t_i-K+1:t_i},
\hat F_{t_i}
\right),
\nonumber\\
\pi_{\hat\phi}
&\in
\arg\min_{\pi_\phi\in\Pi_F}
\widehat{\mathcal R}_\eta(\pi_\phi).
\label{eq:pfc_empirical_risk}
\end{align}

Using
$\mathcal R_\eta(\pi_\phi)
=
\mathcal H_\beta+\mathcal E_\eta(\pi_\phi)$,
we can decompose the population error of the learned policy as
\begin{align}
\mathcal E_\eta(\pi_{\hat\phi})
&=
\mathcal E_\eta(\pi_{\hat\phi})
-
\mathcal E_\eta(\pi_\eta^{\rm P})
+
\mathcal E_\eta(\pi_\eta^{\rm P})
\nonumber\\
&=
\mathcal R_\eta(\pi_{\hat\phi})
-
\mathcal R_\eta(\pi_\eta^{\rm P})
+
\mathcal E_\eta(\pi_\eta^{\rm P})
\nonumber\\
&\leq
\mathcal R_\eta(\pi_{\hat\phi})
-
\mathcal R_\eta(\pi_\eta^{\rm P})
+
\epsilon_{\rm app}^{F}\nonumber\\
&\leq
\Big[
\mathcal R_\eta(\pi_{\hat\phi})
-
\widehat{\mathcal R}_\eta(\pi_{\hat\phi})
\Big]+
\Big[
\widehat{\mathcal R}_\eta(\pi_{\hat\phi})
-
\widehat{\mathcal R}_\eta(\pi_\eta^{\rm P})
\Big]
\nonumber\\
&\quad+
\Big[
\widehat{\mathcal R}_\eta(\pi_\eta^{\rm P})
-
\mathcal R_\eta(\pi_\eta^{\rm P})
\Big]
+
\epsilon_{\rm app}^{F}.
\label{eq:pfc_finite_sample_2}
\end{align}
Since $\pi_{\hat\phi}$ minimizes the empirical risk over $\Pi_F$,
\begin{align}
&\widehat{\mathcal R}_\eta(\pi_{\hat\phi})
-
\widehat{\mathcal R}_\eta(\pi_\eta^{\rm P})
\leq 0,
\nonumber\\
\mathcal E_\eta(\pi_{\hat\phi})
&\leq
2
\sup_{\pi_\phi\in\Pi_F}
\left|
\mathcal R_\eta(\pi_\phi)
-
\widehat{\mathcal R}_\eta(\pi_\phi)
\right|
+
\epsilon_{\rm app}^{F}.
\label{eq:pfc_uniform_reduction}
\end{align}

Let $c_F$ denote an upper bound on the range of the per-sample
negative log-likelihood
$\ell_\phi
=
-\log \pi_\phi
\left(
a_t\mid\tau_{t-K+1:t},\hat F_t
\right)$
over $\phi\in\Pi_F$. When $c_F$ is finite, with probability at least $1-\delta$,
\begin{align}
&\Pr
\Bigg(
\sup_{\pi_\phi\in\Pi_F}
\left|
\mathcal R_\eta(\pi_\phi)
-
\widehat{\mathcal R}_\eta(\pi_\phi)
\right|
>
c_F
\sqrt{
\frac{
\log(2|\Pi_F|/\delta)
}{
2N
}
}
\Bigg)
\leq
\delta,
\nonumber\\
&\sup_{\pi_\phi\in\Pi_F}
\left|
\mathcal R_\eta(\pi_\phi)
-
\widehat{\mathcal R}_\eta(\pi_\phi)
\right|
\leq
c_F
\sqrt{
\frac{
\log(2|\Pi_F|/\delta)
}{
2N
}
}.
\label{eq:pfc_uniform_convergence}
\end{align}
Substituting this result into
~\eqref{eq:pfc_uniform_reduction}, we obtain
\begin{align}
\mathcal E_\eta(\pi_{\hat\phi})
&\leq
\epsilon_{\rm app}^{F}
+
2c_F
\sqrt{
\frac{
\log(2|\Pi_F|/\delta)
}{
2N
}
}.
\label{eq:pfc_finite_sample_error}
\end{align}

Finally, substituting
~\eqref{eq:pfc_finite_sample_error}
into ~\eqref{eq:pfc_performance_reduction}, we have gives
\begin{align}
&
J(\bar{\pi}_\eta^{\rm P})
-
J(\bar{\pi}_{\eta,\hat\phi})
\nonumber\\
&\leq
C_FT^2
\left[
\sqrt{
2\epsilon_{\rm app}^{F}
}
+
\sqrt{
2\epsilon_{\rm app}^{F}
+
4c_F
\sqrt{
\frac{
\log(2|\Pi_F|/\delta)
}{
2N
}
}
}
\right]
\nonumber\\
&\leq
C_FT^2
\left[
2\sqrt{
2\epsilon_{\rm app}^{F}
}
+
2\sqrt{c_F}
\left(
\frac{
\log(2|\Pi_F|/\delta)
}{
2N
}
\right)^{1/4}
\right]
\nonumber\\
&=
\mathcal O
\left(
C_FT^2
\left[
\sqrt{c_F}
\left(
\frac{
\log(|\Pi_F|/\delta)
}{
N
}
\right)^{1/4}
+
\sqrt{
\epsilon_{\rm app}^{F}
}
\right]
\right).
\label{eq:pfc_final_bound}
\end{align}
Proof completed.

\section{Experimental Details}
\subsection{Experimental Tasks and Datasets}

\paragraph{Environments.}
We evaluate DT-JEPA on diverse D4RL benchmarks. For MuJoCo locomotion tasks,
the experiments include four sequential control environments: Hopper, HalfCheetah, Walker2d, and Ant. These
environments cover different levels of dynamical complexity, ranging from
low-dimensional planar locomotion to high-dimensional three-dimensional
quadrupedal control.
\begin{figure}[h]
\centering
\includegraphics[width=1\textwidth]{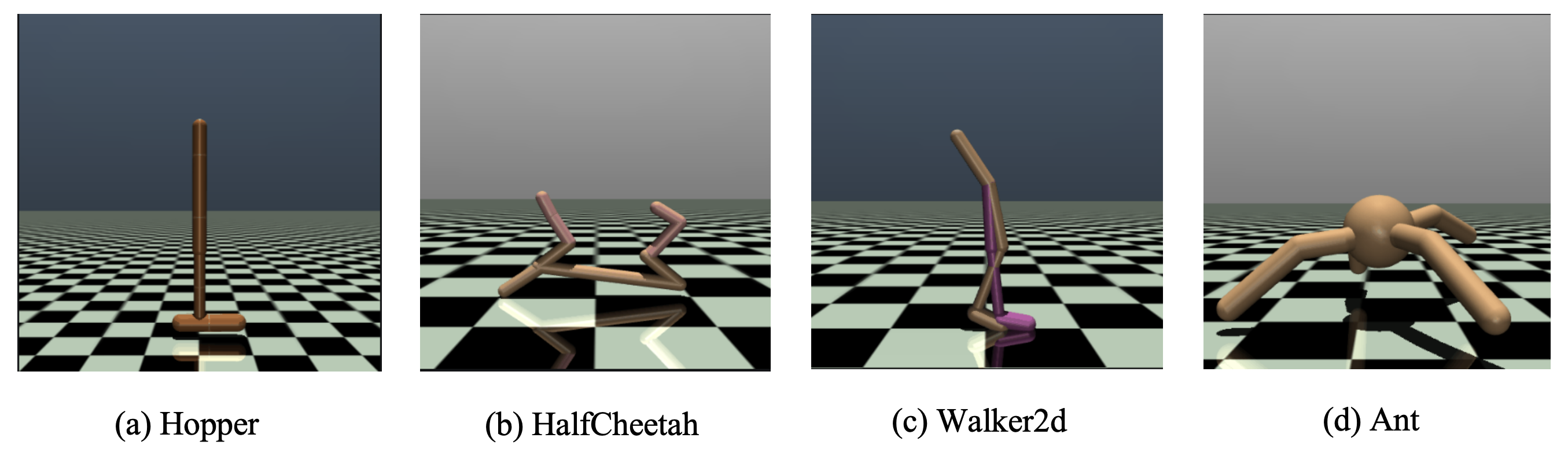}
\caption{
Illustration of four MuJoCo environments utilized in the paper.
}
\label{fig:zero_shot_transfer}
\end{figure}
\begin{itemize}
    \item \textbf{Hopper.}
    Hopper is a planar locomotion environment in which a single-legged robot
    must maintain balance while repeatedly hopping forward. The observation is
    11-dimensional and contains the configurations and velocities of the robot
    body and articulated joints. The action space is 3-dimensional, with each
    action specifying the torque applied to one actuated joint. Due to its small
    action space but sensitive balance dynamics, Hopper provides a relatively
    compact benchmark for evaluating temporal decision making.

    \item \textbf{HalfCheetah.}
    HalfCheetah considers a planar articulated robot whose objective is to
    generate coordinated motions for fast forward locomotion. Its observation
    space contains 17 dimensions describing joint configurations and
    velocities, while the 6-dimensional continuous action controls the torques
    applied to the actuated joints in the front and rear limbs. Compared with
    Hopper, HalfCheetah requires coordination across a larger set of coupled
    joints and admits a wider range of locomotion behaviors.

    \item \textbf{Walker2d.}
    Walker2d is a two-dimensional bipedal locomotion task in which the agent
    controls two articulated legs to move forward while keeping the torso
    upright. The environment has a 17-dimensional observation space and a
    6-dimensional action space. The actions correspond to torques applied to
    the thigh, leg, and foot joints of the two legs. Successful control requires
    both forward progression and stable coordination between the two legs,
    making the task particularly sensitive to errors in long-horizon action
    prediction.

    \item \textbf{Ant.}
    Ant is a three-dimensional quadrupedal locomotion environment with four
    articulated legs and eight actuated joints. The agent must coordinate all
    four legs to produce stable forward motion. In the D4RL MuJoCo setting, the
    observation is 111-dimensional, incorporating joint configurations,
    velocities, and contact-related information, while the action space is
    8-dimensional and specifies the torques applied to the eight controllable
    joints. Its substantially higher state and action complexity makes Ant a
    challenging benchmark for evaluating whether learned representations can
    capture long-horizon and multi-joint dynamics.
\end{itemize}
For Maze2D navigation tasks, we consider 3 environments with increasing
difficulty:  Umaze, Medium, and Large. In Maze2D, the agent controls a
point mass to navigate through a two-dimensional maze toward a fixed goal.
The environment has a 4-dimensional state space describing the position and
velocity of the agent, and a 2-dimensional continuous action space specifying
its acceleration. The reward is determined by the distance between the agent
and the goal. The four environments mainly differ in maze structure and
navigation horizon, providing progressively more challenging settings for
evaluating long-horizon decision making.

\begin{itemize}
    \item \textbf{Umaze.}
    Umaze introduces a compact U-shaped maze that requires the agent to
    navigate around obstacles rather than moving directly toward the goal.
    Compared with Open, successful control requires the policy to account for
    the maze structure and coordinate a sequence of actions over a longer
    navigation horizon.

    \item \textbf{Medium.}
    Medium contains a larger and more complex maze with longer feasible paths
    between the initial position and the goal. The agent must make appropriate
    route choices while maintaining stable motion through multiple corridors,
    increasing the demand for long-term trajectory reasoning.

    \item \textbf{Large.}
    Large is the most challenging Maze2D configuration, featuring the largest
    maze and the longest navigation horizon. Reaching the target requires the
    agent to coordinate decisions over extended trajectories and avoid
    locally attractive but globally ineffective movements, making it a
    challenging benchmark for long-horizon navigation.
\end{itemize}
For AntMaze navigation tasks, we consider three maze configurations:
Umaze, Medium, and Large. AntMaze extends Maze2D by replacing the
point-mass agent with a quadrupedal Ant robot, resulting in a more
challenging control problem that combines locomotion and long-horizon
navigation. The environment has a 27-dimensional state space and an
8-dimensional continuous action space. The three configurations mainly
differ in maze size and navigation complexity, providing progressively
more challenging settings for evaluating long-horizon decision making.
\begin{figure}[h]
\centering
\includegraphics[width=0.8\textwidth]{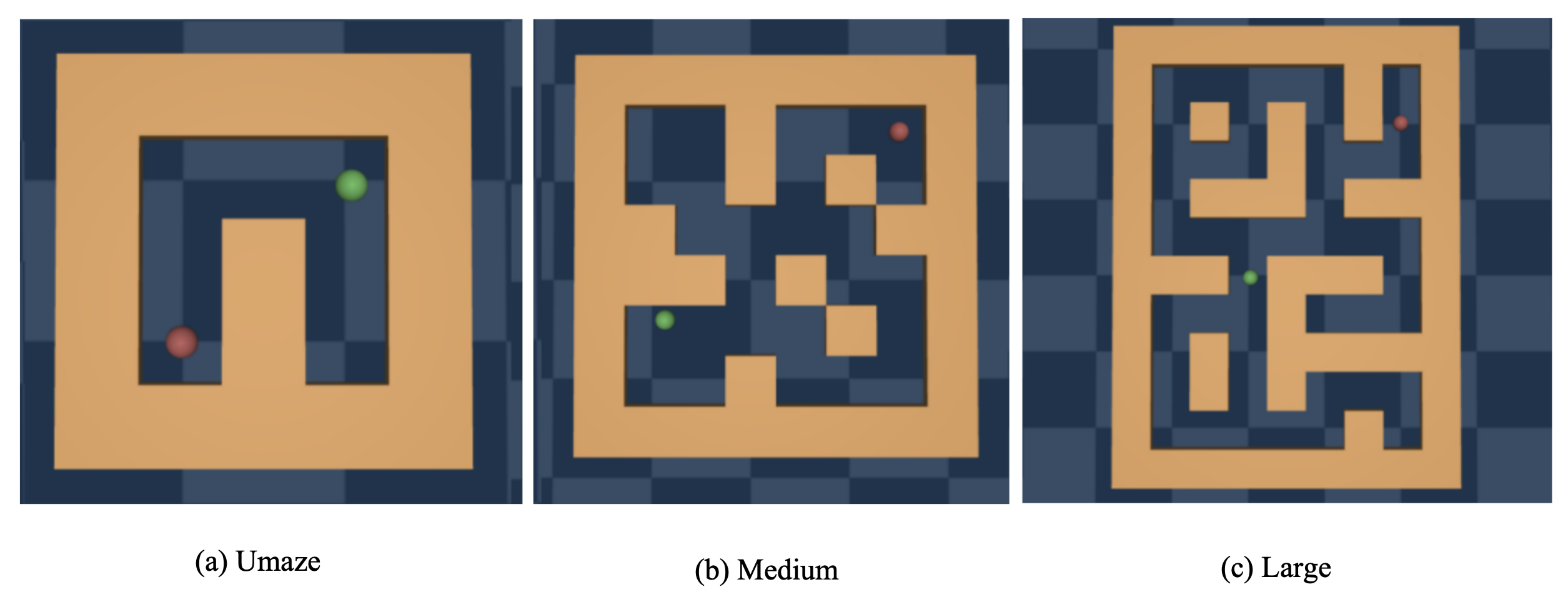}
\caption{
Illustration of four AntMaze environments utilized in the paper.}
\label{fig:zero_shot_transfer}
\end{figure}
\begin{itemize}
    \item \textbf{Umaze.}
    Umaze is the smallest and simplest AntMaze configuration. The agent
    controls the Ant robot to navigate through a compact U-shaped maze
    toward the target location. Due to the relatively short navigation
    distance and simple maze structure, Umaze provides a basic setting for
    evaluating whether the learned policy can combine low-level locomotion
    control with goal-directed navigation.

    \item \textbf{Medium.}
    Medium uses a larger maze with longer navigation paths and more complex
    route choices than Umaze. The agent must maintain stable quadrupedal
    locomotion while navigating through multiple maze corridors toward the
    target. The increased spatial scale requires the policy to capture
    longer-term dependencies between local movements and the overall
    navigation objective.

    \item \textbf{Large.}
    Large is the most challenging of the three AntMaze configurations,
    featuring the largest maze and the longest navigation horizon. Successful
    control requires the agent to coordinate continuous locomotion over a
    long sequence of decisions while following an appropriate route through
    the maze. This setting therefore provides a demanding benchmark for
    evaluating long-horizon planning and the ability to exploit predictive
    representations over extended trajectories.
\end{itemize}

\paragraph{Offline datasets.}
For each locomotion domain, we consider three commonly used D4RL dataset
qualities: \textit{medium}, \textit{medium-replay}, and
\textit{medium-expert}. These datasets differ not only in behavioral quality
but also in the diversity and consistency of the trajectories they contain.

The \textit{medium} dataset is collected using a partially trained policy with
moderate task performance. It therefore mainly contains coherent but
suboptimal behaviors. The \textit{medium-replay} dataset consists of
transitions stored in the replay buffer throughout the training process of the
medium-level policy. As a result, it contains a substantially broader mixture
of behaviors, including early-stage, low-quality, and partially improved
trajectories. The \textit{medium-expert} dataset combines trajectories from
medium and expert policies, producing a heterogeneous dataset that contains
both suboptimal and near-optimal behaviors.

For the AntMaze domain, we consider four commonly used D4RL dataset
configurations: \textit{umaze}, \textit{umaze-diverse},
\textit{medium-play}, and \textit{medium-diverse}. These datasets differ
in both maze complexity and the diversity of navigation behaviors contained
in the offline data.

The \textit{umaze} dataset uses the smallest maze layout with a fixed
starting location and a fixed goal, and therefore contains relatively
consistent navigation behaviors. The \textit{umaze-diverse} dataset uses
the same maze layout but introduces more diverse navigation trajectories
through varying navigation goals and starting conditions. For the larger
Medium maze, \textit{medium-play} is collected using a set of
hand-designed starting and goal locations, producing structured but
heterogeneous navigation behaviors. In contrast,
\textit{medium-diverse} uses more broadly distributed navigation goals,
resulting in a more diverse offline dataset with a wider range of
trajectories. These configurations therefore progressively increase the
difficulty of learning reusable long-horizon navigation behaviors from
offline experience.

For the Maze2D domain, we evaluate on three D4RL environments:
\textit{umaze}, \textit{medium}, and \textit{large}. Unlike AntMaze,
these configurations mainly differ in maze scale and navigation
complexity rather than in the data-collection strategy. The
\textit{umaze} environment contains a compact U-shaped layout and
relatively short navigation paths. The \textit{medium} environment
introduces a larger maze with longer routes and more complex spatial
dependencies, while \textit{large} further increases the maze size and
navigation horizon. The corresponding offline datasets contain continuous
navigation experience toward different goals, making the larger layouts
increasingly challenging for learning long-horizon dependencies and
stitching useful behaviors across trajectories.
\subsection{Baselines}

We provide a more detailed introduction to the baselines used in our experiments in this section.

\begin{itemize}

    \item \textbf{TD3-BC} \citep{fujimoto2021minimalist} is a widely used offline reinforcement learning method built upon Twin Delayed Deep Deterministic Policy Gradient (TD3). It combines value-based policy optimization with a behavior cloning objective that regularizes the learned policy toward actions contained in the offline dataset. This simple modification reduces the tendency of the policy to exploit out-of-distribution actions while still allowing policy improvement through critic guidance. TD3-BC provides a representative non-sequence-modeling offline RL baseline for evaluating whether trajectory-level sequence modeling offers advantages over conventional actor--critic methods.

\item \textbf{TD-JEPA} \citep{bagatella2026tdjepa} is a reward-free
zero-shot RL method based on temporal-difference latent prediction.
It learns state and task encoders together with a policy-conditioned
multi-step predictor from offline reward-free transitions, encouraging
the latent representations to capture long-term policy-dependent
dynamics. The learned predictor approximates successor features in the
latent space and is used to train a family of latent-conditioned policies.
At test time, downstream reward functions are mapped into the learned task
space to retrieve corresponding policies without further policy training.
Since TD-JEPA also exploits JEPA-style latent prediction for reward-free
representation learning, it provides a closely related baseline for
evaluating the predictive representation learning mechanism of DT-JEPA.

    \item \textbf{ODT} \citep{zheng2022online} extends DT to the offline-to-online reinforcement learning setting. It first learns a return-conditioned sequence policy from offline trajectories and subsequently adapts the pretrained policy through limited online environment interactions. In contrast to the deterministic action prediction commonly used in the original DT, ODT employs a stochastic policy and incorporates an entropy-regularized objective to encourage exploration during online fine-tuning. We include ODT as a representative sequence-modeling baseline for evaluating the effectiveness of directly adapting a pretrained DT with online reinforcement learning.

    \item \textbf{QT} \citep{pmlr-v235-hu24c} is a Transformer-based offline
RL method that combines conditional sequence modeling with Q-value
regularization. QT learns an action-value function using an $n$-step
Bellman objective and incorporates the learned Q-values into policy
optimization, while retaining the original DT objective as behavior
regularization. This allows the policy to favor high-value actions while
remaining close to the behavior distribution, thereby improving trajectory
stitching from suboptimal offline data. During inference, QT generates
candidate actions under multiple return conditions and selects actions
according to their estimated Q-values. We include QT to compare DT-JEPA
with a value-guided sequence-modeling approach that explicitly improves
policy optimization and trajectory stitching.
    \item \textbf{Prompt DT} \citep{xu2022prompt} introduces trajectory prompts into DT to improve generalization across tasks. Instead of relying solely on return-to-go and historical state--action sequences, Prompt DT conditions the policy on short trajectory demonstrations collected from the target task. These demonstrations provide task-specific contextual information that allows the Transformer to infer the desired behavior through in-context sequence modeling. Prompt DT therefore serves as an important comparison for evaluating whether explicit task demonstrations are necessary for cross-task adaptation.

    \item \textbf{VDT} \citep{zheng2025value} augments the DT framework with value-based guidance. While standard Decision Transformer mainly learns actions through supervised sequence prediction conditioned on return-to-go, VDT additionally exploits value information to provide a more informative signal for policy optimization. By incorporating value estimation into sequence-based decision making, VDT aims to alleviate the limitations of purely behavior-cloning-style training and improve the quality of generated actions. We use VDT to compare DT-JEPA with a value-guided variant of D.

    \item \textbf{LSDT} \citep{wang2025long} enhances DT by jointly modeling long-term and local dependencies in offline trajectories. Rather than relying on a single temporal representation to capture all historical information, LSDT explicitly considers complementary dependencies at different temporal scales, enabling the policy to better exploit both long-horizon behavioral structure and short-term transition information. This design is particularly relevant for continuous-control problems where current actions may depend simultaneously on recent dynamics and longer-term trajectory context. LSDT is included to assess whether explicitly improving temporal dependency modeling is sufficient to achieve strong generalization.

    \item \textbf{PDT} \citep{xie2023future} is a closely related reward-free sequence-model pretraining method based on future conditioning. During pretraining, PDT encodes a succeeding trajectory segment into a latent future representation and conditions the policy on this latent variable for action prediction, allowing the model to learn diverse behaviors from reward-free and potentially suboptimal trajectories. A future prior is additionally learned to generate latent future representations from the current state. During downstream adaptation, PDT associates future representations with returns through a return prediction model and uses this information to select desirable futures for policy execution. Since PDT explicitly exploits future trajectory information without requiring reward labels during pretraining, it provides one of the most directly related baselines for evaluating DT-JEPA.

\end{itemize}

\section{DT-JEPA Training and Deployment Algorithms}

In this section, we provide the detailed training algorithms of DT-JEPA. As described in the main paper, the overall workflow of DT-JEPA mainly consists of two phases: reward-free offline pretraining and critic-guided online fine-tuning for unseen downstream tasks.

During the offline pretraining phase, DT-JEPA jointly learns predictive future representations and the future-conditioned decision policy from the offline multi-task trajectory dataset $\mathcal D$. Specifically, we sample a historical trajectory segment $\tau_{t-K+1:t}$ and its succeeding future segment $\tau_{t:t+K-1}$ from $\mathcal D$. The historical encoder $f_{\theta}$ maps the observable trajectory history into the latent representation $\mathbf H_t$, while the EMA-updated target encoder $f_{\theta'}$ encodes the succeeding trajectory into the target future representation $\mathbf F_t$. Conditioned on $\mathbf H_t$, the probabilistic predictor $q_{\eta}$ models the distribution of future representations and samples the predicted future latent $\hat{\mathbf F}_t$. The predicted latent is subsequently transformed into feature-wise scaling and shifting parameters and injected into the DT through AdaLN, allowing future-predictive information to modulate the historical representation used for action generation. The JEPA module and DT policy are jointly optimized by minimizing the predictive loss $\mathcal L_{\mathrm{JEPA}}$ and behavior cloning loss $\mathcal L_{\mathrm{BC}}$, while the target encoder is progressively updated through EMA. The detailed pretraining procedure is summarized in Algorithm~1.

During the online fine-tuning phase, the pretrained DT-JEPA policy is adapted to an unseen task $j$ with a new reward objective. The policy first interacts with the new task to collect trajectories and construct the task-specific dataset $\mathcal D^j$, from which a high-return subset $\hat{\mathcal D}^j$ is selected. We then introduce two task-specific critic networks, $Q_{\omega_1^j}^j$ and $Q_{\omega_2^j}^j$, to estimate the long-term value of actions generated by the DT-JEPA policy. The critics are optimized using temporal-difference targets constructed from the minimum value of the two target critics, while EMA target networks are maintained for both the critics and the policy. Based on the learned critics, we further define a value-guided policy objective $\mathcal L_Q^j$ that encourages DT-JEPA to generate actions with higher expected long-term return under the new task. The policy is finally fine-tuned by jointly minimizing $\mathcal L_{\mathrm{BC}}^j$, $\mathcal L_{\mathrm{JEPA}}^j$, and $\mathcal L_Q^j$. The detailed critic-guided online adaptation procedure is provided in Algorithm~2.
\begin{algorithm}[t]
\caption{Reward-Free Pretraining of DT-JEPA}
\label{alg:dtjepa_pretrain}
\begin{algorithmic}[1]

\Require Offline multi-task dataset $\mathcal{D}$, batch size $B$, 
history length $K$, loss weight $\lambda_J$, EMA coefficient $\rho$,
history encoder $f_{\theta}$, target encoder $f_{\theta'}$,
probabilistic predictor $q_{\eta}$, 
DT policy $\phi$

\State Initialize target encoder: $\theta' \leftarrow \theta$

\While{not converged}

    \Statex \hspace{-0.15in} $\triangleright$ \textit{Sample trajectory segments}

    \State Sample
    $\{(\tau_{t-K+1:t}^{(i)},\tau_{t:t+K-1}^{(i)})\}_{i=1}^{B}
    \sim \mathcal{D}$

    \Statex \hspace{-0.15in} $\triangleright$ \textit{Predict future latent representations}

    \For{$i=1,\ldots,B$}

        \State $\mathbf H_t^{(i)}
        \leftarrow
        f_{\theta}
        (\tau_{t-K+1:t}^{(i)})$

        \State $\mathbf F_t^{(i)}
        \leftarrow
        \operatorname{sg}
        \left[
        f_{\theta'}
        (\tau_{t:t+K-1}^{(i)})
        \right]$

        \State
        $q_{\eta}
        (\cdot\mid\mathbf H_t^{(i)})
        =
        \mathcal N
        \left(
        \hat{\boldsymbol\mu}_t^{(i)},
        \operatorname{diag}
        ((\hat{\boldsymbol\sigma}_t^{(i)})^2)
        \right)$

        \State Sample
        $\hat{\mathbf F}_t^{(i)}
        \sim
        q_{\eta}
        (\cdot\mid\mathbf H_t^{(i)})$

    \EndFor

    \Statex \hspace{-0.15in} $\triangleright$ \textit{Compute JEPA prediction loss}

    \State
    $\displaystyle
    \mathcal L_{\mathrm{JEPA}}
    =
    -\frac{1}{B}
    \sum_{i=1}^{B}
    \log
    q_{\eta}
    \left(
    \mathbf F_t^{(i)}
    \mid
    \mathbf H_t^{(i)}
    \right)$

    \Statex \hspace{-0.15in} $\triangleright$ \textit{Future-conditioned action generation}

    \For{$i=1,\ldots,B$}

        \State Obtain DT hidden representation
        $\mathbf h_t^{(i)}
        \leftarrow
        \operatorname{DT}_{\phi}
        (\tau_{t-K+1:t}^{(i)})$

        \State
        $\boldsymbol\gamma_t^{(i)}
        \leftarrow
        \operatorname{MLP}_{\gamma}
        (\hat{\mathbf F}_t^{(i)})$,
        \quad
        $\boldsymbol\beta_t^{(i)}
        \leftarrow
        \operatorname{MLP}_{\beta}
        (\hat{\mathbf F}_t^{(i)})$

        \State
        $\tilde{\mathbf h}_t^{(i)}
        \leftarrow
        (1+\boldsymbol\gamma_t^{(i)})
        \odot
        \operatorname{LayerNorm}
        (\mathbf h_t^{(i)})
        +
        \boldsymbol\beta_t^{(i)}$

        \State Predict actions
        $\hat a_m^{(i)}
        \leftarrow
        \phi
        (\tau_{t-K+1:t}^{(i)},
        \hat{\mathbf F}_t^{(i)})$

    \EndFor

    \Statex \hspace{-0.15in} $\triangleright$ \textit{Compute behavior cloning loss}

    \State
    $\displaystyle
    \mathcal L_{\mathrm{BC}}
    =
    \frac{1}{B}
    \sum_{i=1}^{B}
    \sum_{m=t-K+1}^{t}
    \left\|
    a_m^{(i)}-\hat a_m^{(i)}
    \right\|_2^2$

    \State Update $\phi,\theta,\eta$ to minimize
    $\mathcal L_{\mathrm{BC}}
    +\lambda_J\mathcal L_{\mathrm{JEPA}}$

    \State Update target encoder via EMA:
    $\theta'
    \leftarrow
    \rho\theta'
    +(1-\rho)\theta$

\EndWhile

\State \Return pretrained DT-JEPA policy $\phi$, $f_{\theta}$, and $q_{\eta}$.

\end{algorithmic}
\end{algorithm}
\begin{algorithm}[t]
\caption{Critic-Guided Online Fine-Tuning of DT-JEPA}
\label{alg:dtjepa_finetune}
\begin{algorithmic}[1]

\Require Pretrained DT-JEPA policy $\phi$, JEPA predictor $q_{\eta}$,
critic loss weight $\lambda_Q$, JEPA loss weight $\lambda_J$,
discount factor $\gamma$, EMA coefficient $\rho$


\State Initialize task-specific dataset $\mathcal D^j$

\State Initialize critics
$Q_{\omega_1^j}^{j}$ and $Q_{\omega_2^j}^{j}$

\State Initialize target networks:
$\bar{\omega}_1^j\leftarrow\omega_1^j$,
$\bar{\omega}_2^j\leftarrow\omega_2^j$,
$\bar{\phi}^j\leftarrow\phi^j$

\While{not converged}

    \Statex \hspace{-0.15in} $\triangleright$ \textit{Collect task-specific trajectories}

    \State Add newly collected trajectories to $\mathcal D^j$

    \State Construct high-return subset
    $\hat{\mathcal D}^{j}\subseteq\mathcal D^j$

    \State Sample
    $\{\tau_{t-K+1:t}^{j,(i)}\}_{i=1}^{B}
    \sim\hat{\mathcal D}^{j}$

    \Statex \hspace{-0.15in} $\triangleright$ \textit{Update task-specific critics}

    \For{$i=1,\ldots,B$}

        \State Generate target action
        $\hat a_t^{j,(i)}
        \sim
        \bar{\phi}
        (\tau_{t-K+1:t}^{j,(i)})$

        \State For $m=t-K+1,\ldots,t-1$, compute
        \[
        \hat Q_m^{j,(i)}
        =
        \sum_{k=m}^{t-1}
        \gamma^{k-m}r_k^{j,(i)}
        +
        \gamma^{t-m}
        \min_{d\in\{1,2\}}
        Q_{\bar{\omega}_d^j}^{j}
        \left(
        s_t^{j,(i)},
        \hat a_t^{j,(i)}
        \right)
        \]

    \EndFor

    \For{$d\in\{1,2\}$}

        \State
        $\displaystyle
        \mathcal L_{\mathrm{critic}}^{j}
        (\omega_d^j)
        =
        \frac{1}{B}
        \sum_{i=1}^{B}
        \sum_{m=t-K+1}^{t-1}
        \left(
        Q_{\omega_d^j}^{j}
        (s_m^{j,(i)},a_m^{j,(i)})
        -
        \hat Q_m^{j,(i)}
        \right)^2$

        \State Update $\omega_d^j$ to minimize
        $\mathcal L_{\mathrm{critic}}^{j}(\omega_d^j)$

    \EndFor

    \Statex \hspace{-0.15in} $\triangleright$ \textit{Critic-guided policy optimization}

    \State Generate DT-JEPA actions
    $\hat a_m^{j,(i)}$
    under the current policy $\phi^j$, $f_{\theta}^j$, and $p_{\eta}^j$.

    \State
    $\displaystyle
    \mathcal L_Q^j
    =
    -
    \frac{1}{B}
    \sum_{i=1}^{B}
    \sum_{m=t-K+1}^{t-1}
    Q_{\omega_1^j}^{j}
    \left(
    s_m^{j,(i)},
    \hat a_m^{j,(i)}
    \right)$

    \State Update $\phi^j,\theta^j,\eta^j$ to minimize
    \[
    \mathcal L_{\mathrm{BC}}^j
    +
    \lambda_J\mathcal L_{\mathrm{JEPA}}^j
    +
    \lambda_Q\mathcal L_Q^j
    \]

    \State Update target critics via EMA:
    \[
    \bar{\omega}_d^j
    \leftarrow
    \rho\bar{\omega}_d^j
    +(1-\rho)\omega_d^j,
    \qquad d\in\{1,2\}
    \]

    \State Update target policy via EMA:
    $\bar{\phi}^j
    \leftarrow
    \rho\bar{\phi}^j
    +(1-\rho)\phi^j$

\EndWhile

\State \Return adapted DT-JEPA policy $\phi^j$, $f_{\theta}^j$, and $q_{\eta}^j$. 

\end{algorithmic}
\end{algorithm}
\section{Hyperparameters}
Table
\ref{tab:dtjepa_hyperparameters} summarizes the common hyperparameters
used across all experiments. Unless otherwise specified, the same
configuration is used for all environments.

\begin{table}[!t]
\centering
\scriptsize
\setlength{\tabcolsep}{5pt}
\renewcommand{\arraystretch}{0.9}
\resizebox{0.6\linewidth}{!}{
\begin{tabular}{lc}
\toprule
\textbf{Hyperparameter} & \textbf{Value} \\
\midrule

JEPA encoder architecture & lightweight Transformer \\
JEPA encoder layers & 2 \\
future latent dimension $d_F$ & 64 \\
context length $K$ & 20 \\
EMA coefficient & 0.99 \\

\midrule

DT Transformer layers & 4 \\
DT attention heads & 4 \\
DT embedding dimension & 512 \\
dropout & 0.1 \\
batch size & 256 \\
learning rate & $1\times10^{-4}$ \\
weight decay & $1\times10^{-3}$ \\
gradient norm clip & 0.25 \\

\midrule

critic layer & $2$ \\
critic width & $256$ \\
discount factor $\gamma$ & 0.99 \\
critic learning rate & $1\times10^{-3}$ \\

target update coefficient & 0.005 \\

\bottomrule
\end{tabular}
}
\caption{Common hyperparameters used for DT-JEPA.}
\label{tab:dtjepa_hyperparameters}
\end{table}
\section{Construction of Unseen Reward Tasks}
\label{app:unseen_reward_tasks}

To evaluate the ability of DT-JEPA to generalize beyond the reward
objectives observed during pretraining, we construct a set of unseen
downstream tasks by modifying the reward functions of the four MuJoCo
locomotion environments: Hopper, HalfCheetah, Walker2d, and Ant.
Across these tasks, the state space, action space, transition dynamics,
and termination conditions of each environment remain unchanged; only
the reward function is modified. Therefore, the resulting tasks share
the same underlying dynamics while requiring different behavioral
objectives.

For each environment, we consider three downstream objectives:
\textit{run forward}, \textit{run backward}, and \textit{jump}.
These reward functions are not used during reward-free pretraining and
are introduced only during downstream adaptation.

Let $v_x$ denote the velocity of the agent along the $x$-axis,
$\bm a$ denote the action vector, $z$ denote the current torso height,
and $z_{\mathrm{init}}$ denote the initial torso height of the episode.
For Ant, $c_{\mathrm{contact}}$ denotes the contact cost.

\paragraph{HalfCheetah.}
Following~\citet{yu2021conservative}, the forward, backward, and jump
tasks are defined as
\begin{align}
r_{\mathrm{HC}}^{\mathrm{forward}}(s,\bm a)
&=
\max\{v_x,3\}
-0.1\|\bm a\|_2^2,
\\
r_{\mathrm{HC}}^{\mathrm{backward}}(s,\bm a)
&=
-\max\{v_x,3\}
-0.1\|\bm a\|_2^2,
\\
r_{\mathrm{HC}}^{\mathrm{jump}}(s,\bm a)
&=
-0.1\|\bm a\|_2^2
+15(z-z_{\mathrm{init}}).
\end{align}
The first two objectives encourage locomotion along opposite horizontal
directions, whereas the jump objective removes the horizontal velocity
reward and instead encourages upward body displacement.

\paragraph{Walker2d.}
For Walker2d, the corresponding reward functions are
\begin{align}
r_{\mathrm{W}}^{\mathrm{forward}}(s,\bm a)
&=
v_x
-0.001\|\bm a\|_2^2,
\\
r_{\mathrm{W}}^{\mathrm{backward}}(s,\bm a)
&=
-v_x
-0.001\|\bm a\|_2^2,
\\
r_{\mathrm{W}}^{\mathrm{jump}}(s,\bm a)
&=
-|v_x|
-0.001\|\bm a\|_2^2
+10(z-z_{\mathrm{init}}).
\end{align}
For the jump task, the horizontal-velocity penalty discourages the agent
from obtaining high reward simply through horizontal locomotion, while
the height term encourages vertical movement.

\paragraph{Ant.}
For the Ant environment, we additionally retain the contact penalty in
all three objectives:
\begin{align}
r_{\mathrm{A}}^{\mathrm{forward}}(s,\bm a)
&=
v_x
-0.5\|\bm a\|_2^2
-0.005c_{\mathrm{contact}},
\\
r_{\mathrm{A}}^{\mathrm{backward}}(s,\bm a)
&=
-v_x
-0.5\|\bm a\|_2^2
-0.005c_{\mathrm{contact}},
\\
r_{\mathrm{A}}^{\mathrm{jump}}(s,\bm a)
&=
-|v_x|
-0.5\|\bm a\|_2^2
-0.005c_{\mathrm{contact}}
+10(z-z_{\mathrm{init}}).
\end{align}
The contact and control penalties regularize the behavior of the
high-dimensional quadruped, while the velocity and height terms define
the task-specific objective.

\paragraph{Hopper.}
Hopper tasks following the same design principle as Walker2d,
which has a similar planar locomotion setting:
\begin{align}
r_{\mathrm{H}}^{\mathrm{forward}}(s,\bm a)
&=
v_x
-0.001\|\bm a\|_2^2,
\\
r_{\mathrm{H}}^{\mathrm{backward}}(s,\bm a)
&=
-v_x
-0.001\|\bm a\|_2^2,
\\
r_{\mathrm{H}}^{\mathrm{jump}}(s,\bm a)
&=
-|v_x|
-0.001\|\bm a\|_2^2
+10(z-z_{\mathrm{init}}).
\end{align}

\section{Addtional Results}
This section reports additional experiments, as well as detailed numerical results for plots in the
main part of the paper.
\subsection{Effect of predicted future representations on the action distributions.}
\begin{figure}[h]
\centering
\includegraphics[width=1\textwidth]{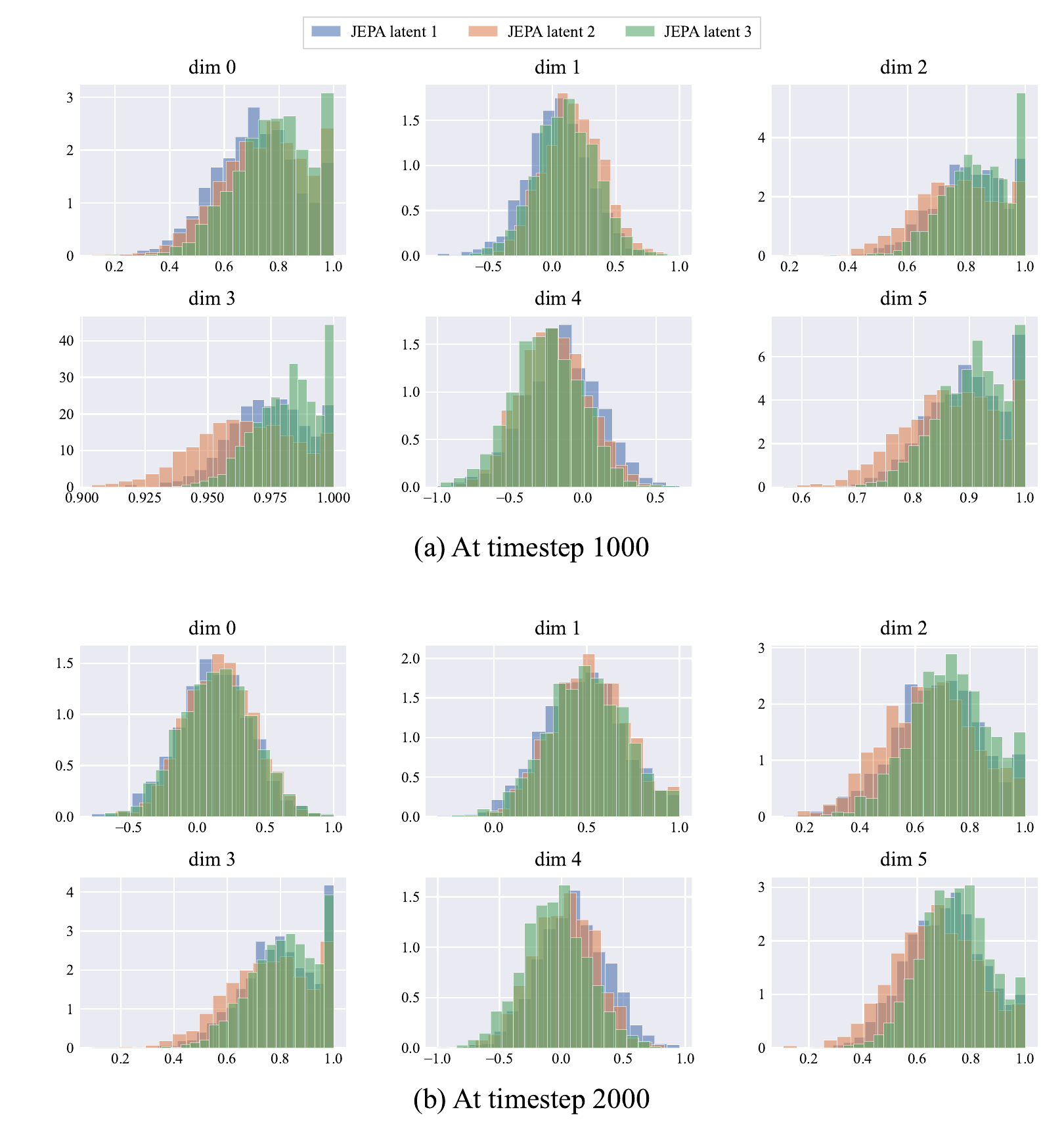}
\caption{
Action distributions produced by DT-JEPA under different latent conditions of an episode. Each subplot corresponds to one dimension of the continuous action space. The horizontal axis denotes the action value, while the vertical axis represents the normalized probability density of the histogram, computed as $\hat{p}_i = N_i/(N\Delta a)$, where $N_i$ is the number of action samples falling into the $i$-th bin, $N$ is the total number of sampled actions, and $\Delta a$ is the bin width.
}
\label{fig:zero_shot_transfer}
\end{figure}
To examine whether the JEPA-predicted future representations effectively
participate in decision making, we visualize the action distributions induced
by different sampled future latents. We conduct this analysis on the Walker2d task using a DT-JEPA agent
pretrained on the corresponding offline dataset, and visualize the action
distributions under different future representations sampled from the
probabilistic JEPA predictor. Given the same historical context, we sample three future representations from the probabilistic
JEPA predictor and use them to condition the DT policy through AdaLN. As shown
in Figure~\ref{fig:zero_shot_transfer}, different JEPA latents induce clearly distinguishable action
distributions across multiple action dimensions at both timesteps 1000 and
2000. Since the observable history is kept unchanged,
the resulting variation can be attributed to the predicted future
representation, confirming that $\hat{\mathbf F}_t$ is actively used by the
policy rather than being ignored or collapsed into an ineffective conditioning
signal.

More importantly, the differences are structured across action dimensions
rather than appearing as uniform perturbations. This suggests that different
samples from the predictive distribution encode distinct possible future
evolutions, which are translated by AdaLN into different control preferences.
In this sense, the probabilistic JEPA predictor does not merely provide an
additional stochastic input, but induces a family of future-conditioned action
distributions under the same observed history. These observations provide evidence that the
predicted latent future serves as a meaningful decision variable linking
future-dynamics prediction to action generation in DT-JEPA.
\subsection{Effect of predicted future representations on the attention score.}
\begin{figure}[h]
\centering
\includegraphics[width=1\textwidth]{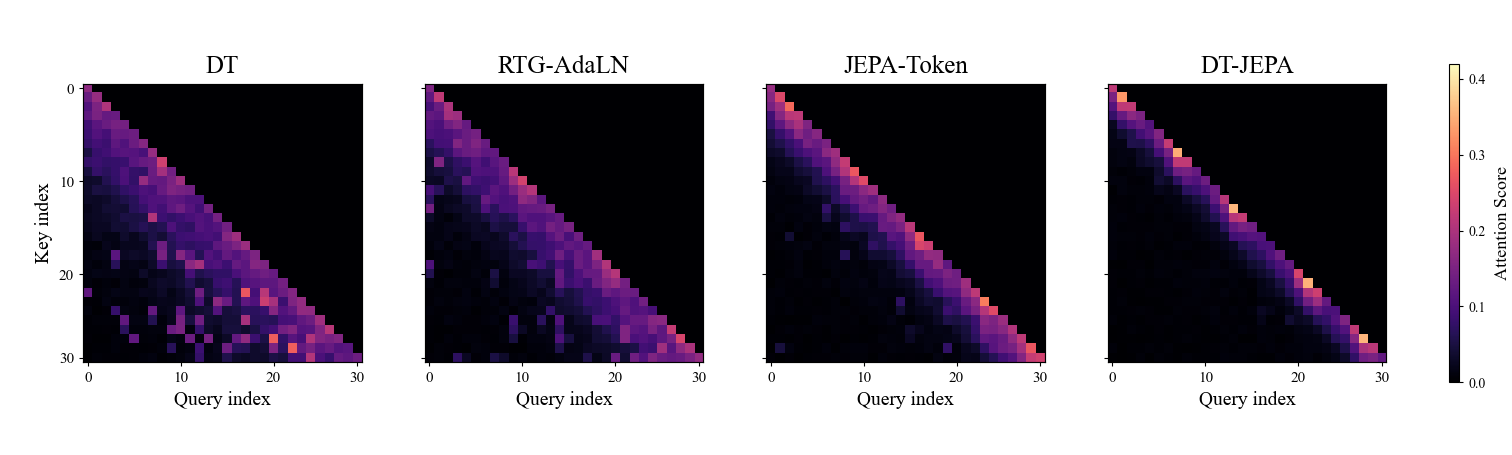}
\caption{
\textbf{Attention score visualization on HalfCheetah-medium-replay.}
Compared with DT, RTG-AdaLN, and JEPA-Token, DT-JEPA exhibits a more
concentrated attention pattern around the main diagonal. Since HalfCheetah
is dominated by strong local temporal dependencies, this diagonal structure
indicates that the model places greater emphasis on temporally adjacent and
dynamically relevant trajectory information. The comparison further highlights
the complementary effects of predictive future conditioning and AdaLN modulation.
}
\label{fig:attention_visualization}
\end{figure}

To further understand how different future-conditioning strategies affect historical
information utilization, we visualize the attention scores of four variants in
Figure~\ref{fig:attention_visualization}. Here, \textbf{DT} adopts the conventional
RTG token as the conditioning signal; \textbf{RTG-AdaLN} retains RTG but injects it
through AdaLN; \textbf{JEPA-Token} replaces RTG with the predicted future
representation $\hat{\mathbf F}_t$ while preserving token-based conditioning; and
\textbf{DT-JEPA} combines predictive future representations with AdaLN modulation.

A key observation is that the attention patterns become progressively more concentrated
around the main diagonal from DT to DT-JEPA. This phenomenon is particularly meaningful
for HalfCheetah, whose dynamics exhibit a strong local Markov structure: the state and
action at the current step are primarily determined by temporally nearby transitions,
while distant historical tokens are generally less informative for immediate action
prediction. Therefore, an attention pattern concentrated around the diagonal indicates
that the model assigns larger weights to temporally adjacent and dynamically relevant
trajectory information rather than distributing attention broadly over the entire
history.

The original DT exhibits a relatively diffuse attention pattern, suggesting that the
RTG-conditioned Transformer does not sharply distinguish between locally relevant and
less informative historical tokens. Introducing AdaLN with the same RTG condition
already increases the locality of attention, as shown by RTG-AdaLN, indicating that
feature-wise modulation provides a more direct mechanism for conditioning the
self-attention computation. Replacing RTG with the predicted future representation
further strengthens the diagonal structure in JEPA-Token. This suggests that
$\hat{\mathbf F}_t$, which encodes information about the expected future dynamics,
provides a more informative signal for identifying which recent trajectory elements are
most relevant to the current decision.

DT-JEPA exhibits the clearest and most concentrated diagonal attention pattern among all
variants. This result indicates that predictive future conditioning and AdaLN modulation
play complementary roles: the predicted future representation provides information
about the likely subsequent dynamics, while AdaLN injects this information directly into
the historical feature representations, allowing the Transformer to more selectively
emphasize temporally local transitions that are most relevant under the underlying
Markov dynamics.
\subsection{Hyperparameter Sensitivity}

\subsubsection{Sensitivity to $\lambda_J$ During Training}
We further investigate the sensitivity of DT-JEPA to the training-stage
weight $\lambda_J$ of the JEPA objective by varying it over
$\{0.001, 0.01, 0.1, 1, 10\}$. For the MuJoCo locomotion tasks, all
experiments are conducted on the \textit{medium} datasets. For navigation
tasks, AntMaze is evaluated on \textit{umaze}, while Maze2D includes the
\textit{umaze}, \textit{medium}, and \textit{large} configurations.

\begin{figure*}[t]
    \centering
    \includegraphics[width=\textwidth]{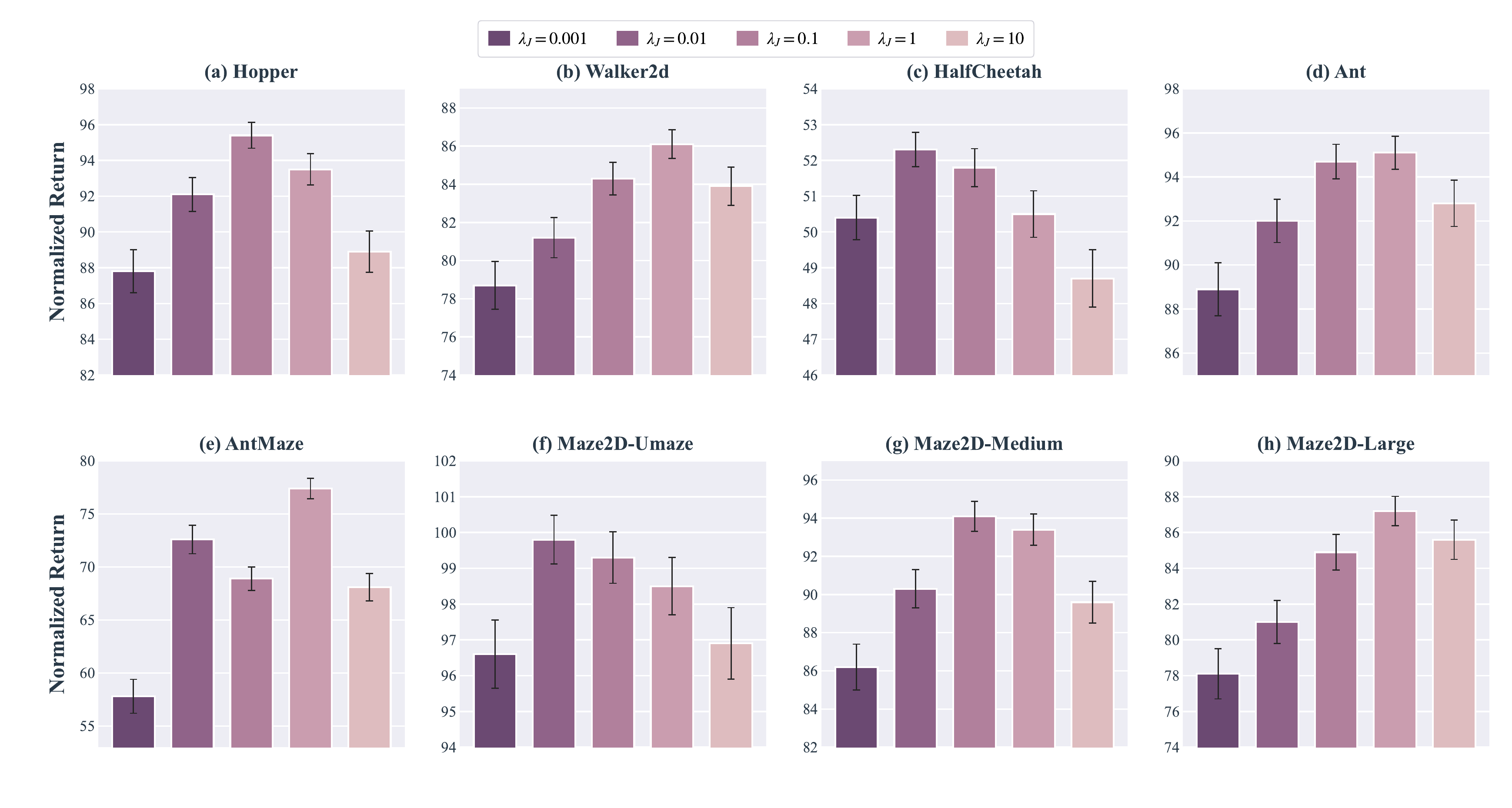}
    \caption{
    Sensitivity analysis of the training-stage JEPA loss weight $\lambda_J$ across different control and navigation environments.
    We vary $\lambda_J$ over $\{0.001, 0.01, 0.1, 1, 10\}$ and report the normalized return.
    Error bars indicate the standard deviation over multiple random seeds.
    }
    \label{fig:lambda_sensitivity}
\end{figure*}

As shown in Fig.~\ref{fig:lambda_sensitivity}, the preferred value of
$\lambda_J$ varies across environments. Hopper performs best around
$\lambda_J=0.1$, whereas Walker2d, AntMaze, and Maze2D-Large favor a
relatively stronger JEPA objective around $\lambda_J=1$. In contrast,
HalfCheetah and Maze2D-Umaze achieve better performance with smaller
values of $\lambda_J$. Despite these differences, DT-JEPA remains
relatively stable over a moderate range of $\lambda_J$, indicating that
its performance does not depend on a narrowly tuned value.

The observed variation also reflects the different roles of predictive
representation learning across tasks. A small $\lambda_J$ causes training
to be dominated by behavior cloning, limiting the amount of future
dynamics information preserved in the learned representation. Increasing
$\lambda_J$ strengthens the predictive objective and can be particularly
beneficial in tasks requiring longer-term temporal reasoning, as suggested
by the preference for larger values in AntMaze and Maze2D-Large.
However, an excessively large $\lambda_J$ shifts the optimization too far
toward latent prediction and may reduce the emphasis on accurate action
generation. Simpler or more locally determined tasks therefore do not
necessarily benefit from stronger JEPA supervision. 
\subsubsection{Sensitivity to $\lambda_J'$ and $\lambda_Q$ During Deployment}
\begin{figure}[t]
    \centering
    \includegraphics[width=1\textwidth]{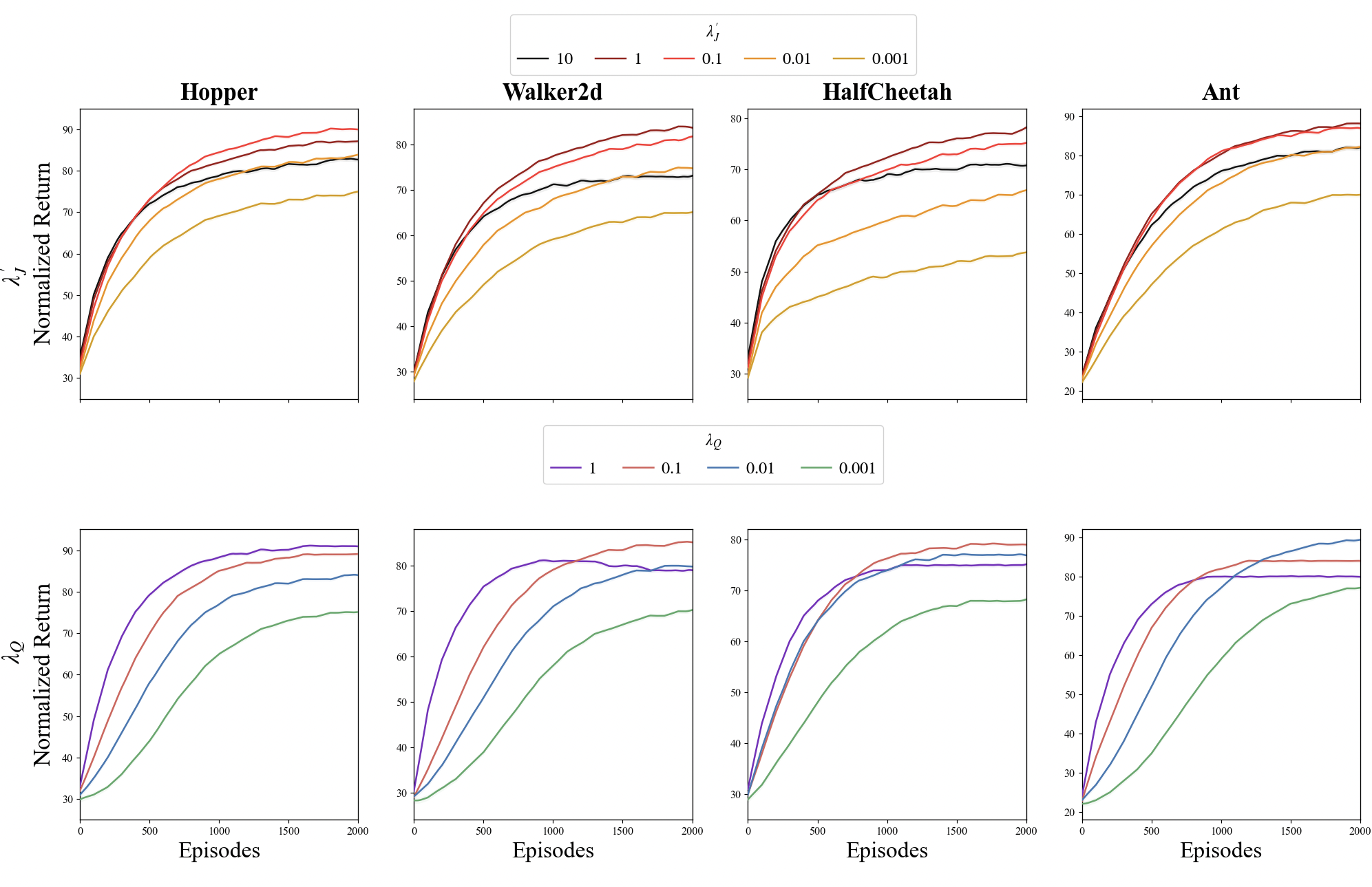}
    \caption{
    \textbf{Sensitivity analysis of $\lambda_J$ and $\lambda_Q$ on
    four MuJoCo Tasks.}
    The upper panel varies the weight of the JEPA predictive objective
    $\lambda_J$, while the below panel varies the critic-guidance coefficient
    $\lambda_Q$. 
    }
    \label{fig:hyperparameter_sensitivity}
\end{figure}

We further analyze the sensitivity of DT-JEPA to the two training-stage
coefficients $\lambda_J$ and $\lambda_Q$ on four MuJoCo environments.
As shown in Fig.~\ref{fig:hyperparameter_sensitivity}, the preferred
hyperparameter values vary across environments, indicating that the
balance between predictive representation learning and value-guided
policy optimization is task-dependent.

For $\lambda_J'$, moderate values generally provide better performance,
while excessively small values weaken the contribution of JEPA-based
future representation learning and overly large values may place too
much emphasis on the auxiliary predictive objective. In particular,
Hopper favors $\lambda_J=0.1$, whereas Walker2d, HalfCheetah, and Ant
perform well with $\lambda_J$ around $1$ or $0.1$.

For $\lambda_Q$, the main difference is reflected in the adaptation
dynamics. A relatively large $\lambda_Q$ can accelerate early-stage
adaptation in some environments, while a smaller value may provide more
stable improvements at later stages. For example, Walker2d exhibits a
clear late-stage advantage with $\lambda_Q=0.1$, whereas Ant benefits
more from the milder critic guidance induced by $\lambda_Q=0.01$.
Overall, DT-JEPA shows stable performance over a moderate range of both
hyperparameters, while the optimal trade-off depends on the environment.

\subsection{Analysis on Trajectory Stitching Ability}
\label{sec:trajectory_stitching}

\begin{table*}[!t]
\centering
\renewcommand{\arraystretch}{1.02}
\setlength{\tabcolsep}{3.6pt}

\resizebox{0.88\textwidth}{!}{
\begin{tabular}{lccccccccc}
\toprule
\textbf{Dataset}
& \textbf{QT}
& \textbf{Prompt DT}
& \textbf{VDT}
& \textbf{LSDT}
& \textbf{PDT}
& \textbf{DT-JEPA} \\
\midrule

Hopper-m
& $\underline{84.8}$
& $78.3$
& $77.8$
& $77.9$
& $77.6$
& $\mathbf{86.6}$ \\

Hopper-m-r
& $\mathbf{76.8}$
& $60.2$
& $63.4$
& $67.9$
& $65.2$
& $\underline{75.9}$ \\

Hopper-m-e
& $98.7$
& $91.5$
& $94.2$
& $\underline{99.4}$
& $96.8$
& $\mathbf{101.2}$ \\

\midrule

Walker2d-m
& $80.5$
& $73.2$
& $73.5$
& $\underline{80.8}$
& $73.4$
& $\mathbf{83.2}$ \\

Walker2d-m-r
& $\underline{74.3}$
& $64.9$
& $68.0$
& $72.6$
& $60.7$
& $\mathbf{76.8}$ \\

Walker2d-m-e
& $106.2$
& $99.3$
& $102.0$
& $\mathbf{109.0}$
& $103.7$
& $\underline{108.6}$ \\

\midrule

Halfcheetah-m
& $\underline{48.1}$
& $43.0$
& $43.9$
& $43.2$
& $36.5$
& $\mathbf{50.3}$ \\

Halfcheetah-m-r
& $44.8$
& $39.1$
& $38.7$
& $\underline{45.5}$
& $30.2$
& $\mathbf{47.0}$ \\

Halfcheetah-m-e
& $92.0$
& $86.2$
& $89.1$
& $\underline{92.7}$
& $90.4$
& $\mathbf{94.1}$ \\

\midrule

Ant-m
& $\underline{88.8}$
& $77.5$
& $80.2$
& $80.9$
& $79.8$
& $\mathbf{92.0}$ \\

Ant-m-r
& $73.5$
& $64.7$
& $69.0$
& $\underline{74.8}$
& $67.2$
& $\mathbf{76.4}$ \\

Ant-m-e
& $108.5$
& $101.7$
& $105.4$
& $\underline{109.8}$
& $106.2$
& $\mathbf{111.4}$ \\

\midrule

Antmaze-u
& $\underline{60.5}$
& $44.2$
& $51.0$
& $57.2$
& $54.6$
& $\mathbf{63.8}$ \\

Antmaze-u-d
& $\underline{49.2}$
& $32.0$
& $39.8$
& $46.1$
& $43.8$
& $\mathbf{52.4}$ \\

Antmaze-m-p
& $14.6$
& $20.1$
& $29.2$
& $\underline{35.8}$
& $33.4$
& $\mathbf{41.5}$ \\

Antmaze-m-d
& $\underline{30.9}$
& $13.7$
& $23.4$
& $28.2$
& $26.7$
& $\mathbf{33.1}$ \\

\midrule

Maze2d-u
& $\mathbf{93.5}$
& $75.0$
& $82.1$
& $8.4$
& $85.9$
& $\underline{92.6}$ \\

Maze2d-m
& $\underline{82.6}$
& $65.2$
& $55.2$
& $14.7$
& $77.1$
& $\mathbf{85.4}$ \\

Maze2d-l
& $\underline{74.8}$
& $52.6$
& $65.5$
& $10.2$
& $69.4$
& $\mathbf{77.6}$ \\

\bottomrule
\end{tabular}
}

\caption{
Trajectory stitching ability on degraded D4RL locomotion and navigation
datasets. For each dataset, the top $40\%$ highest-return complete
trajectories are removed before training, resulting in offline datasets
that contain more fragmented and suboptimal behaviors.
All methods are trained using the same degraded datasets without
additional environment interaction.
The best result is shown in \textbf{bold}, while the second-best result
is \underline{underlined}.
}
\label{tab:trajectory_stitching}

\end{table*}
As shown in Table~\ref{tab:trajectory_stitching}, removing the top
high-return trajectories makes the offline datasets substantially more
fragmented and reduces the availability of complete high-quality behaviors.
Under this setting, DT-JEPA maintains strong performance across most
locomotion and navigation tasks, indicating that its policy does not rely
solely on imitating complete successful trajectories. Instead, the learned
future representations appear to provide a useful predictive structure for
connecting locally effective behavior segments that may originate from
different trajectories.

This advantage is particularly evident on the long-horizon navigation
tasks. DT-JEPA performs consistently well across the AntMaze variants and
the more challenging Maze2D configurations, where successful behavior often
requires combining useful sub-trajectories over extended horizons. This
suggests that predicting future latent dynamics helps the policy reason
beyond the immediate local transition and identify behavior fragments that
are compatible with a desirable future evolution. In contrast, methods that
mainly rely on sequence imitation or fixed future conditioning are generally
more affected when coherent high-return trajectories are removed, since the
offline data provide weaker direct supervision for complete decision
sequences.
QT remains highly competitive on several datasets, which is consistent with
the benefit of value-based trajectory stitching: its learned Q-function can
favor locally high-value actions even when they are observed in different
trajectories. DT-JEPA, however, shows a more consistent advantage across the
degraded datasets, suggesting that predictive future representations provide
a complementary way to recover long-horizon behavioral structure without
explicitly relying on value-based action selection. 

\subsection{Sensitivity to the JEPA Predictor Depth}
\label{sec:predictor_depth}

\begin{table}[t]
\centering
\renewcommand{\arraystretch}{1.00}
\setlength{\tabcolsep}{3.5pt}

\caption{
Sensitivity analysis of the JEPA predictor depth $L_P$ on D4RL
locomotion and navigation tasks.
$L_P$ denotes the number of Transformer blocks in the latent-space
JEPA predictor.
Results are reported as normalized returns over five random seeds.
The best result is shown in \textbf{bold}, while the second-best
result is \underline{underlined}.
}
\label{tab:predictor_depth}

\scriptsize
\resizebox{0.92\linewidth}{!}{
\begin{tabular}{lcccc}
\toprule
\textbf{Dataset}
& \textbf{$L_P=1$}
& \textbf{$L_P=2$}
& \textbf{$L_P=3$}
& \textbf{$L_P=4$}
\\
\midrule

Hopper-m
& $89.6 \pm 2.5$
& $94.1 \pm 1.9$
& $\mathbf{95.4 \pm 1.7}$
& $\underline{94.6 \pm 1.8}$
\\

Hopper-m-r
& $90.4 \pm 3.4$
& $96.2 \pm 2.8$
& $\mathbf{98.3 \pm 2.1}$
& $\underline{96.9 \pm 2.6}$
\\

Hopper-m-e
& $100.6 \pm 2.0$
& $104.4 \pm 1.5$
& $\mathbf{105.2 \pm 1.3}$
& $\underline{104.7 \pm 1.4}$
\\

\midrule

Walker2d-m
& $79.6 \pm 3.6$
& $84.4 \pm 3.1$
& $\underline{86.1 \pm 2.8}$
& $\mathbf{86.5 \pm 2.8}$
\\

Walker2d-m-r
& $86.3 \pm 4.2$
& $92.1 \pm 3.6$
& $\underline{93.1 \pm 2.9}$
& $\mathbf{94.4 \pm 2.3}$
\\

Walker2d-m-e
& $107.1 \pm 2.2$
& $111.3 \pm 1.8$
& $\underline{112.3 \pm 1.5}$
& $\mathbf{112.3 \pm 1.2}$
\\

\midrule

Halfcheetah-m
& $49.7 \pm 1.3$
& $\mathbf{52.3 \pm 0.8}$
& $\underline{52.3 \pm 0.9}$
& $51.7 \pm 1.0$
\\

Halfcheetah-m-r
& $46.5 \pm 1.7$
& $\mathbf{49.8 \pm 1.0}$
& $\underline{49.8 \pm 1.2}$
& $49.3 \pm 1.3$
\\

Halfcheetah-m-e
& $93.4 \pm 1.4$
& $\underline{95.9 \pm 1.1}$
& $\mathbf{96.2 \pm 1.0}$
& $95.8 \pm 1.5$
\\

\midrule

Ant-m
& $89.0 \pm 3.2$
& $94.1 \pm 2.7$
& $\mathbf{95.1 \pm 2.4}$
& $\underline{94.4 \pm 2.5}$
\\

Ant-m-r
& $71.8 \pm 4.6$
& $78.2 \pm 4.0$
& $\mathbf{80.6 \pm 3.6}$
& $\underline{79.5 \pm 3.8}$
\\

Ant-m-e
& $109.5 \pm 2.3$
& $113.7 \pm 1.9$
& $\mathbf{114.5 \pm 1.7}$
& $\underline{114.1 \pm 1.8}$
\\

\midrule

Antmaze-u
& $66.6 \pm 4.2$
& $74.2 \pm 3.6$
& $\mathbf{77.4 \pm 3.1}$
& $\underline{76.4 \pm 3.4}$
\\

Antmaze-u-d
& $82.9 \pm 5.0$
& $91.1 \pm 4.3$
& $\mathbf{94.9 \pm 2.9}$
& $\underline{93.6 \pm 4.1}$
\\

Antmaze-m-p
& $39.7 \pm 5.3$
& $48.3 \pm 4.6$
& $\mathbf{52.9 \pm 4.0}$
& $\underline{51.4 \pm 4.3}$
\\

Antmaze-m-d
& $59.9 \pm 5.6$
& $68.8 \pm 4.8$
& $\mathbf{73.9 \pm 3.1}$
& $\underline{72.2 \pm 4.6}$
\\

\midrule

Maze2d-u
& $91.1 \pm 3.3$
& $97.1 \pm 2.7$
& $\underline{99.8 \pm 2.4}$
& $\mathbf{99.8 \pm 2.1}$
\\

Maze2d-m
& $84.6 \pm 3.9$
& $91.0 \pm 3.2$
& $\underline{93.5 \pm 2.6}$
& $\mathbf{94.1 \pm 2.9}$
\\

Maze2d-l
& $76.8 \pm 4.4$
& $83.6 \pm 3.7$
& $\underline{84.5 \pm 2.3}$
& $\mathbf{87.2 \pm 3.3}$
\\

\midrule

Average
& $79.7$
& $85.3$
& $\mathbf{87.1}$
& $\underline{86.8}$
\\

\bottomrule
\end{tabular}
}
\end{table}

We further investigate the sensitivity of DT-JEPA to the capacity of
the latent predictor by varying the number of Transformer blocks
$L_P\in\{1,2,3,4\}$ while keeping the JEPA encoder, policy backbone,
and all other training configurations fixed.
As shown in Table~\ref{tab:predictor_depth}, the predictor depth has a
clear influence on downstream performance.
Increasing $L_P$ from one to two blocks improves the average normalized
return from $79.7$ to $85.3$, while using three blocks further increases
it to $87.4$.
This suggests that a very shallow predictor provides insufficient
capacity for modeling the conditional evolution of future latent
representations.
In contrast, increasing the depth from three to four blocks does not
yield further overall improvement, with the average return slightly
decreasing from $87.4$ to $86.9$.
Therefore, the benefit of increasing predictor capacity gradually
saturates once a moderate depth is reached.

The effect of predictor depth also varies noticeably across task
families.
For the locomotion benchmarks, performance improves relatively smoothly
from one to three blocks.
Hopper and Ant consistently favor a three-block predictor, whereas
HalfCheetah reaches comparable or slightly better performance with only
two blocks.
This behavior is consistent with the relatively smooth dynamics of
HalfCheetah, for which additional predictor capacity brings limited
benefit.
Walker2d exhibits a slightly different pattern, where four blocks
provide marginal improvements over three blocks.
However, the differences between $L_P=3$ and $L_P=4$ remain small,
suggesting that the additional capacity is not essential for these
locomotion tasks.

The dependence on predictor depth becomes more pronounced on the
navigation benchmarks.
Averaged over AntMaze and Maze2D, increasing the predictor depth from
one to three blocks produces a substantially larger improvement than
that observed on the locomotion tasks.
In particular, AntMaze benefits consistently from increasing $L_P$ up
to three, with clear gains on the more challenging diverse and
medium-scale tasks.
This suggests that reasoning over longer-horizon and more multimodal
future transitions requires a sufficiently expressive latent predictor.
Maze2D shows a similar improvement as the predictor becomes deeper,
although four blocks yield only marginal gains over three.
These results indicate that additional predictor capacity is especially
useful when the future evolution is more temporally extended or admits
multiple possible behaviors.

\subsection{Sensitivity to the Sequence Length}
\label{sec:k_sensitivity}

\begin{table}[t]
\centering
\renewcommand{\arraystretch}{1.00}
\setlength{\tabcolsep}{3.2pt}

\caption{
Sensitivity analysis of the sequence length $K$ on D4RL locomotion
and navigation tasks. Results are reported as normalized returns over
five random seeds. The best result is shown in \textbf{bold}, while
the second-best result is \underline{underlined}.
}
\label{tab:k_sensitivity}

\scriptsize
\resizebox{0.96\linewidth}{!}{
\begin{tabular}{lccccc}
\toprule
\textbf{Dataset}
& \textbf{$K=5$}
& \textbf{$K=10$}
& \textbf{$K=20$}
& \textbf{$K=30$}
& \textbf{$K=40$}
\\
\midrule

Hopper-m
& $89.8 \pm 2.8$
& $93.7 \pm 2.1$
& $\mathbf{95.4 \pm 1.7}$
& $\underline{94.9 \pm 1.9}$
& $94.0 \pm 2.2$
\\

Hopper-m-r
& $89.2 \pm 3.9$
& $94.7 \pm 3.0$
& $\mathbf{98.3 \pm 2.1}$
& $\underline{97.5 \pm 2.4}$
& $96.5 \pm 2.7$
\\

Hopper-m-e
& $101.4 \pm 2.2$
& $103.8 \pm 1.8$
& $\mathbf{105.2 \pm 1.3}$
& $\underline{104.8 \pm 1.5}$
& $104.3 \pm 1.7$
\\

\midrule

Walker2d-m
& $80.2 \pm 4.0$
& $84.6 \pm 3.3$
& $\mathbf{86.1 \pm 2.8}$
& $\underline{85.8 \pm 3.0}$
& $85.2 \pm 3.3$
\\

Walker2d-m-r
& $86.8 \pm 3.9$
& $91.8 \pm 3.1$
& $\mathbf{94.4 \pm 2.3}$
& $\underline{93.6 \pm 2.6}$
& $92.7 \pm 3.0$
\\

Walker2d-m-e
& $108.2 \pm 2.3$
& $110.9 \pm 1.9$
& $\underline{112.3 \pm 1.5}$
& $\mathbf{112.5 \pm 1.6}$
& $111.6 \pm 1.9$
\\

\midrule

Halfcheetah-m
& $49.8 \pm 1.5$
& $51.4 \pm 1.2$
& $\mathbf{52.3 \pm 0.9}$
& $\underline{51.9 \pm 1.0}$
& $51.6 \pm 1.2$
\\

Halfcheetah-m-r
& $46.9 \pm 2.0$
& $48.6 \pm 1.6$
& $\mathbf{49.8 \pm 1.2}$
& $\underline{49.5 \pm 1.4}$
& $49.0 \pm 1.6$
\\

Halfcheetah-m-e
& $93.7 \pm 1.8$
& $95.1 \pm 1.4$
& $\mathbf{96.2 \pm 1.0}$
& $\underline{95.9 \pm 1.1}$
& $95.4 \pm 1.3$
\\

\midrule

Ant-m
& $88.7 \pm 4.0$
& $92.8 \pm 3.2$
& $\mathbf{95.1 \pm 2.4}$
& $\underline{94.7 \pm 2.6}$
& $93.9 \pm 3.0$
\\

Ant-m-r
& $72.4 \pm 5.2$
& $77.9 \pm 4.4$
& $\mathbf{80.6 \pm 3.6}$
& $\underline{79.9 \pm 3.9}$
& $78.8 \pm 4.3$
\\

Ant-m-e
& $110.2 \pm 2.6$
& $112.8 \pm 2.1$
& $\mathbf{114.5 \pm 1.7}$
& $\underline{114.2 \pm 1.8}$
& $113.5 \pm 2.1$
\\

\midrule

Antmaze-u
& $63.1 \pm 5.0$
& $71.8 \pm 4.1$
& $\underline{77.4 \pm 3.1}$
& $\mathbf{77.8 \pm 3.3}$
& $75.6 \pm 3.8$
\\

Antmaze-u-d
& $79.8 \pm 4.7$
& $88.9 \pm 3.9$
& $\underline{94.9 \pm 2.9}$
& $\mathbf{95.2 \pm 3.1}$
& $92.6 \pm 3.6$
\\

Antmaze-m-p
& $36.8 \pm 5.9$
& $46.0 \pm 5.0$
& $\underline{52.9 \pm 4.0}$
& $\mathbf{53.3 \pm 4.2}$
& $49.7 \pm 4.8$
\\

Antmaze-m-d
& $58.2 \pm 5.0$
& $66.8 \pm 4.2$
& $\underline{73.9 \pm 3.1}$
& $\mathbf{74.3 \pm 3.3}$
& $70.5 \pm 3.9$
\\

\midrule

Maze2d-u
& $88.0 \pm 4.1$
& $95.0 \pm 3.3$
& $\mathbf{99.8 \pm 2.4}$
& $\underline{99.4 \pm 2.6}$
& $97.2 \pm 3.0$
\\

Maze2d-m
& $79.5 \pm 4.8$
& $88.7 \pm 4.0$
& $\underline{94.1 \pm 2.9}$
& $\mathbf{94.4 \pm 3.1}$
& $91.8 \pm 3.6$
\\

Maze2d-l
& $69.4 \pm 5.3$
& $80.6 \pm 4.5$
& $\underline{87.2 \pm 3.3}$
& $\mathbf{87.7 \pm 3.5}$
& $84.3 \pm 4.1$
\\

\midrule

Average
& $78.5$
& $84.0$
& $\mathbf{87.4}$
& $\underline{87.2}$
& $85.7$
\\

\bottomrule
\end{tabular}
}
\end{table}
We further study the sensitivity of DT-JEPA to the sequence length $K$,
which determines the temporal range available for predictive representation
learning and sequence-based policy modeling. As shown in
Table~\ref{tab:k_sensitivity}, increasing $K$ from 5 to 20 consistently
improves performance across almost all environments. In particular, the
average normalized return increases substantially as the context grows from
$K=5$ to $K=10$ and further to $K=20$. This trend suggests that very short
sequences provide insufficient temporal information for DT-JEPA to infer
reliable future dynamics and establish useful dependencies between historical
observations and subsequent behaviors.

The benefit of increasing $K$, however, gradually saturates once the context
reaches a moderate length. Among the twelve MuJoCo locomotion datasets,
$K=20$ achieves the best performance on almost all tasks, while $K=30$
provides only marginal additional benefit on Walker2d-medium-expert.
This indicates that locomotion control mainly requires a moderate temporal
window: once sufficient recent dynamics are captured, extending the sequence
further contributes little additional decision-relevant information.
Conversely, the navigation environments exhibit a somewhat different pattern.
Most AntMaze tasks and the larger Maze2D configurations slightly favor
$K=30$, suggesting that longer contexts can be beneficial when decisions
depend on longer-term spatial and temporal dependencies. This effect becomes
more apparent as the navigation horizon and maze complexity increase, whereas
the simpler Maze2D-Umaze task already reaches its best performance at
$K=20$.
Importantly, further increasing the sequence length to $K=40$ does not lead
to additional improvement and instead causes a consistent performance drop.
This suggests that excessively long contexts may introduce redundant or
weakly relevant historical information, making it more difficult for the
predictive encoder and policy Transformer to identify the most useful
temporal dependencies. Longer sequences also enlarge the temporal span over
which future representations must summarize the trajectory, potentially
making the prediction target more heterogeneous and difficult to model.
Therefore, simply providing more history is not always beneficial; an
appropriate context length should balance sufficient temporal coverage with
the concentration of decision-relevant information.

\subsection{Additional Ablation Results}
\begin{table}[t]
\centering
\renewcommand{\arraystretch}{0.96}
\setlength{\tabcolsep}{3.2pt}

\caption{
Additional ablation experiment.
PDT-AdaLN applies AdaLN-based conditioning to PDT,
PDT-CG augments PDT with the same critic-guided adaptation used in
DT-JEPA, and PDT-AdaLN-CG combines both modifications.
All methods are evaluated on the same unseen reward objectives across
Hopper, Walker2d, HalfCheetah, and Ant as in
Table~\ref{tab:dtjepa_ablation}.
}
\label{tab:full_controlled_ablation}

\scriptsize
\resizebox{0.72\linewidth}{!}{
\begin{tabular}{lccccc}
\toprule
\textbf{Method}
& \textbf{Hopper}
& \textbf{Walker2d}
& \textbf{HalfCheetah}
& \textbf{Ant}
& \textbf{Avg.}
\\
\midrule

DT-RTG-AdaLN
& 71.0
& 67.8
& 71.9
& 64.6
& 68.8
\\

PDT
& 77.2
& 73.7
& 80.5
& 72.4
& 76.0
\\

PDT-AdaLN
& 78.4
& 75.0
& 81.3
& 73.6
& 77.1
\\

PDT-CG
& 80.2
& 77.2
& 82.8
& 75.2
& 78.9
\\

PDT-AdaLN-CG
& \underline{81.5}
& \underline{78.1}
& \underline{83.9}
& \underline{75.5}
& \underline{79.5}
\\

\textbf{DT-JEPA}
& \textbf{89.4}
& \textbf{85.6}
& \textbf{91.2}
& \textbf{83.1}
& \textbf{87.3}
\\

\bottomrule
\end{tabular}
}
\end{table}

Table~\ref{tab:full_controlled_ablation} further provides a controlled
comparison on the same unseen downstream tasks as
Table~\ref{tab:dtjepa_ablation}, aiming to determine whether the
generalization advantage of DT-JEPA can be mainly attributed to
AdaLN-based conditioning or critic-guided adaptation. Compared with
RTG-based conditioning, PDT already achieves substantially stronger
performance, suggesting that incorporating future-oriented information
provides a more informative basis for adapting the policy to reward
objectives that are not observed during pretraining.

Applying AdaLN to PDT leads to a consistent but relatively modest
improvement. This trend is consistent with the main ablation results,
where replacing conventional RTG conditioning with AdaLN alone only
provides limited gains. These observations indicate that AdaLN improves
how the conditioning information interacts with the policy, but does not
fundamentally determine the quality of the conditioning signal itself.
In other words, a more flexible modulation mechanism is beneficial only
when the underlying representation contains sufficiently informative
task-relevant structure.

A more pronounced improvement is observed when critic-guided adaptation
is introduced to PDT. This shows that downstream value information is
important for aligning a pretrained future-conditioned policy with an
unseen reward objective. Combining critic guidance with AdaLN further
improves PDT across the four environments, indicating that the two
mechanisms play complementary roles: AdaLN facilitates the utilization
of conditioning information inside the policy, whereas critic guidance
provides reward-oriented supervision that steers the policy toward
desirable behaviors during adaptation.

Nevertheless, a clear performance gap remains between PDT-AdaLN-CG and
the full DT-JEPA on all evaluated unseen tasks. More importantly, when
considered together with Table~\ref{tab:dtjepa_ablation},
PDT-AdaLN-CG still performs below DT-JEPA without critic guidance.
This comparison is particularly informative because the strengthened PDT
baseline has already been equipped with both AdaLN and the same
critic-guided adaptation mechanism used by DT-JEPA. Therefore, the
remaining advantage cannot be explained solely by the modulation
architecture or downstream critic supervision. Instead, it points to the
importance of the future conditioning representation itself.

The key distinction is that DT-JEPA does not rely only on a generic
future embedding as the conditioning variable. It explicitly predicts
future latent representations from the observed trajectory history and
models their conditional distribution through JEPA learning. As a result,
the conditioning signal is directly grounded in the current behavioral
context and captures predictive information about how the trajectory may
evolve. Such history-dependent future information can provide a more
transferable decision prior when the downstream reward objective changes,
which helps explain why DT-JEPA retains an advantage even after the
alternative baseline is equipped with comparable conditioning and
adaptation mechanisms.

\end{document}

%% file: math_commands.tex
\usepackage{amsmath,amsfonts,bm}

\def\eqref#1{equation~\ref{#1}}

\def\1{\bm{1}}

\DeclareMathAlphabet{\mathsfit}{\encodingdefault}{\sfdefault}{m}{sl}
\SetMathAlphabet{\mathsfit}{bold}{\encodingdefault}{\sfdefault}{bx}{n}

